\documentclass[runningheads]{llncs}

\usepackage[mobile]{eccv}

\usepackage{eccvabbrv}

\usepackage{graphicx}
\usepackage{booktabs}

\usepackage[accsupp]{axessibility}  

\usepackage{longtable}

\usepackage{array}     
\usepackage{xltabular} 

\usepackage{tcolorbox}
\usepackage{xcolor}

\usepackage{pdflscape} 
\usepackage[accsupp]{axessibility}  

\usepackage{hyperref}

\usepackage{orcidlink}
\usepackage[utf8]{inputenc}
\usepackage{booktabs} 
\usepackage{multirow} 
\usepackage{makecell} 
\usepackage{amssymb} 

\begin{document}

\title{CollectionLoRA: Collecting 50 Effects in 1 LoRA via Multi-Teacher On-Policy Distillation} 
\titlerunning{CollectionLoRA}
\author{Fangtai Wu\inst{1,2} \and
Hailong Guo\inst{2} \and
Shijie Huang\inst{2} \and
Jiayi Song\inst{2,3} \and
Yubo Huang\inst{2} \and
Mushui Liu\inst{1} \and
Zhao Wang\inst{1} \and
Yunlong Yu\inst{1}$^*$ \and
Jiaming Liu\inst{2}$^{*\dagger}$ \and
Ruihua Huang\inst{2}}

\authorrunning{F. Wu et al.}
\vspace{-0.2cm}
\institute{
Zhejiang University \and
Qwen Applications Business Group of Alibaba \and
Xi'an Jiaotong University\\
\email{wft@zju.edu.cn, liujiaming.ljl@alibaba-inc.com}\\
$^*$ Corresponding authors. $^\dagger$ Project lead.
}






\maketitle

\vspace{-0.7cm}

\begin{figure*}[h]
    \centering
    \includegraphics[width=\textwidth]{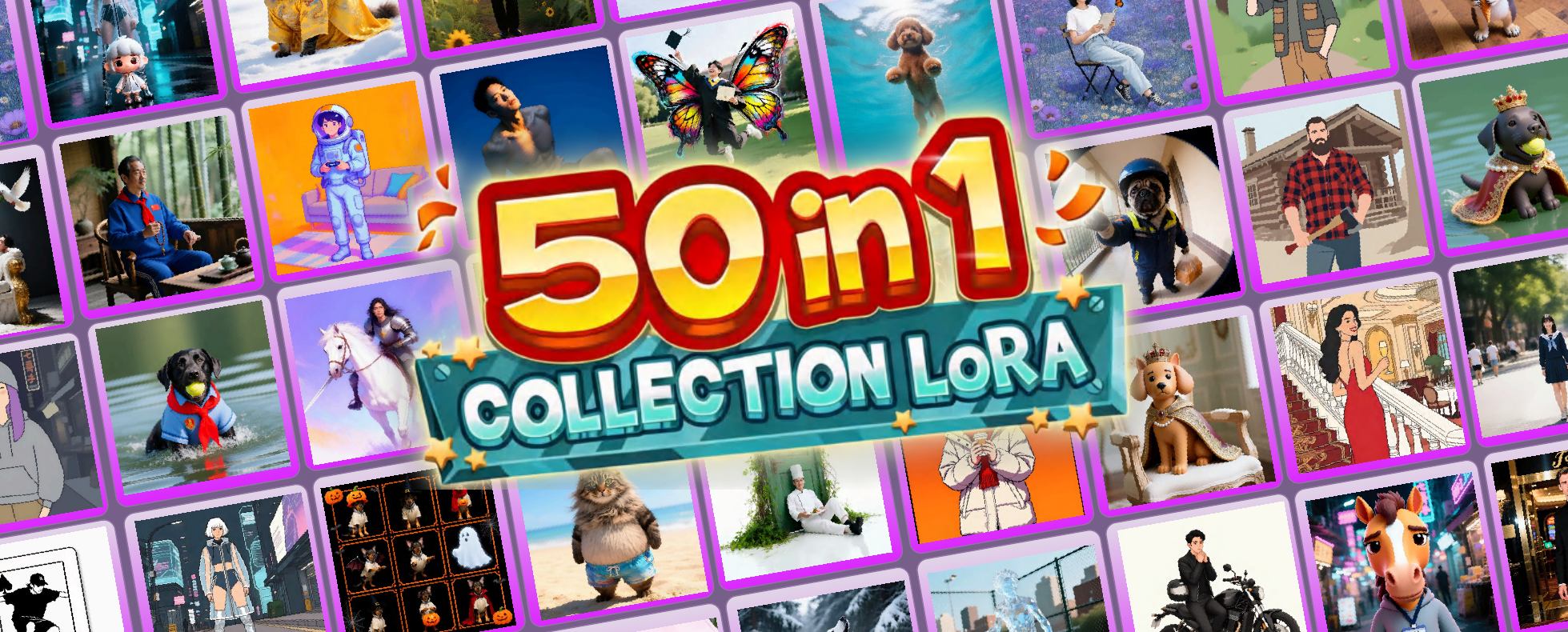}
    \caption{We propose CollectionLoRA, a multi-teacher distillation framework capable of consolidating diverse effects and few-step inference capabilities into a single LoRA.}
    \label{fig:eccv_cover}
\end{figure*}

\vspace{-1.0cm}

\begin{abstract}

Customized image editing aims to equip pre-trained diffusion models with specific visual effects using limited paired data, typically via Low-Rank Adaptation (LoRA). As the number of desired effects grows, storing and dynamically loading numerous these effect LoRAs significantly increases deployment overhead. Furthermore, current pipelines typically cascade these effect LoRAs with acceleration modules for fast generation, which triggers severe parameter interference and results in concept bleeding and style degradation.
\textbf{We propose CollectionLoRA, a multi-teacher on-policy distillation framework} capable of distilling the concepts of up to 50 different effect LoRAs along with few-step generation capabilities into a single LoRA. This fundamentally resolves the feature interference issue and significantly reduces deployment costs.  Specifically, the method introduces (i) a Probabilistic Dual-Stream Routing mechanism that enables the model to randomly switch between data sources during training, effectively enhancing its generalization in unseen scenarios; (ii) an Asymmetric Orthogonal Prompting strategy to achieve concept isolation within the prompt space; (iii) a Coarse-to-Fine Distillation Objective to mitigate the distribution gap between the teacher and student models.
Extensive evaluations show that CollectionLoRA distills all customized effects and few-step generation into a single LoRA, reducing deployment overhead while achieving concept fidelity comparable to or better than independently trained teacher models.
\end{abstract}

\vspace{-0.8cm}

\section{Introduction}
\label{sec:intro}
\begin{figure*}[!t]
    \centering
    \includegraphics[width=\linewidth]{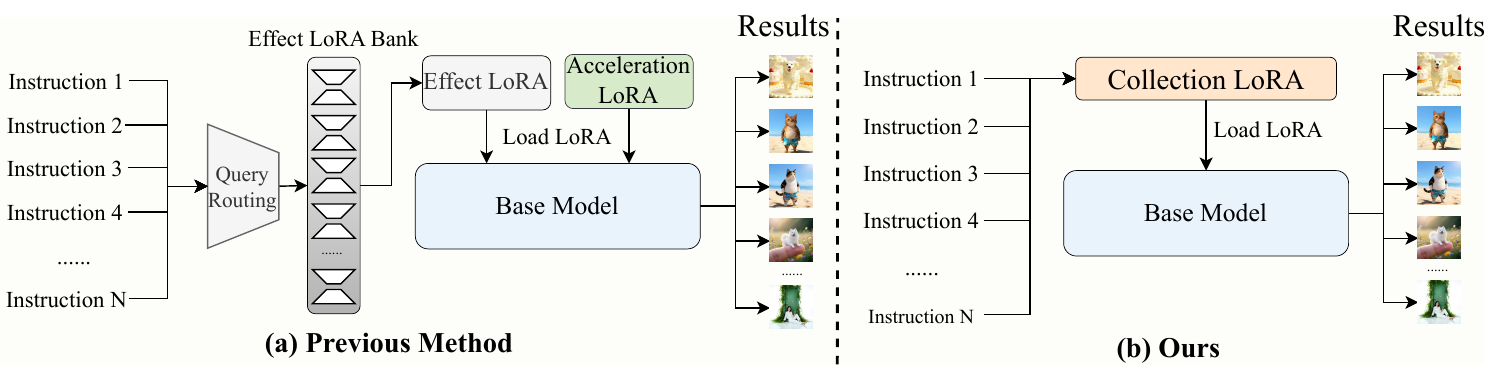}
    \captionof{figure}{Comparison between conventional multi-LoRA pipelines and the proposed CollectionLoRA. \textbf{Conventional}: Training and deploying task-specific LoRA weights for each concept, which are sequentially composed with an acceleration LoRA during inference. \textbf{CollectionLoRA}: Consolidating the acceleration prior and all target concepts into a single unified module through multi-teacher distillation.}
    \label{fig:intro}
    \vspace{-0.4cm}
\end{figure*}

Recently, diffusion models~\cite{flux2024,labs2025flux,flux-2-2025,qwenimage,sd1,esser2024scaling,peebles2023scalable} have revolutionized the field of image editing, enabling unprecedented fine-grained control and high-quality content modification. For customized image editing~\cite{mou2025dreamo,gal2022image,kumari2023multi,wu2025dcoardeepconceptinjection, Photodoodle,ye2023ip,zhang2023addingconditionalcontroltexttoimage,guo2025any2anytryon,zhang2025easycontrol,xie2023omnicontrol,huang2024incontextloradiffusiontransformers,liu2025llm4gen,she2025mosaic,liu2025tfcustom}, the community typically trains specific effect LoRAs~\cite{LoRA,huang2024incontextloradiffusiontransformers,OmniConsistency} using limited paired data and cascades them with acceleration LoRA during inference to achieve rapid, few-step generation. 
However, scaling this paradigm in practice exposes three bottlenecks as illustrated in Fig.~\ref{fig:intro}(a): (i) \textbf{Storage costs.} Deploying all effect LoRAs imposes substantial storage overhead on individual devices. (ii) \textbf{Routing latency and errors.} Retrieving and dynamically loading specific LoRAs from the LoRA bank introduces inference latency and the risk of routing mismatches. (iii) \textbf{LoRA conflicts.} Linearly combining effect and acceleration LoRAs disrupts the original feature manifolds, inevitably causing concept bleeding and style degradation.

To fundamentally address deployment challenges, we aim to consolidate diverse visual effects into a single LoRA. While the concept of distilling knowledge from various domain experts into a unified student model has achieved remarkable success in the realm of Large Language Models (LLMs)~\cite{yang2025qwen3technicalreport,coreteam2026mimov2flashtechnicalreport,deepseek_v4_pro}, it remains largely unexplored within the field of diffusion models. In this work, we treat individual effect LoRAs as distinct visual experts and build our framework upon Distribution Matching Distillation (DMD)~\cite{dmd1,dmd2}.  However, directly applying standard DMD to a multi-teacher setting poses severe challenges. First, initializing the student from the base model creates a massive distribution gap with the experts, leading to distribution collapse during training. Second, consolidating diverse concepts within a shared parameter space with limited data not only causes conflicts between concepts but also degrades the model's generalization ability. To address these challenges, we propose \textbf{CollectionLoRA, a Multi-Teacher On-Policy Distillation framework for diffusion models}. CollectionLoRA stabilizes multi-teacher distillation via three key components: (i) We design a Probabilistic Dual-Stream Routing (PDSR) mechanism that dynamically introduces unlabeled general-domain data as regularization to preserve the model's generalization ability. (ii)  We introduce an Asymmetric Orthogonal Prompting (AOP) strategy. By assigning original prompts to teachers and VLM-rewritten prompts with orthogonal trigger words to the student, it isolates distinct concepts in the latent space and eliminates manual tuning. (iii) Finally, we propose a Coarse-to-Fine Distillation Objective (C2F-DO) to bridge the distribution gap between the student and experts. It combines flow matching~\cite{lipman2023flowmatchinggenerativemodeling} to prevent distribution collapse with Target Simulation (TS) to restore realistic fine details.
In summary, the main contributions are threefold:
\vspace{-0.1cm}
\begin{itemize}
    \item \textbf{New Deployment Paradigm.} We are the first to systematically identify three critical bottlenecks in the conventional multi-LoRA deployment pipeline—storage overhead, routing latency, and parameter conflicts—and propose CollectionLoRA, a multi-teacher on-policy distillation framework that consolidates diverse visual effects and few-step generation into a single LoRA, fundamentally resolving these issues.

    \item \textbf{Effective Multi-Teacher On-Policy Distillation Framework.} To address the 
    unique challenges of multi-teacher distillation, we introduce 
    three key components: \textit{Probabilistic Dual-Stream Routing} (PDSR) for regularization and generalization preservation, \textit{Asymmetric 
    Orthogonal Prompting} (AOP) for concept isolation in the prompt space, and a 
    \textit{Coarse-to-Fine Distillation Objective} (C2F-DO) that synergizes 
    trajectory anchoring with distribution matching to stabilize optimization and restore high-frequency details.
    \item \textbf{Superior Performance and Scalability.} Extensive experiments on EffectBench demonstrate that CollectionLoRA distills \textbf{50 visual effects 
    and few-step generation} into a single LoRA, surpassing independent single-task teachers in concept fidelity while reducing deployment costs. Our framework further \textbf{scales to 180 effects}, reducing deployment overhead to 
    \textbf{0.5\%} of the conventional paradigm without catastrophic quality degradation. Beyond individual effects, we discover an \textbf{zero-shot effect composition} capability, where multiple effects can be seamlessly combined 
    at inference time without any additional training.
\end{itemize}


\section{Related Work}
\label{sec:related_work}

\subsection{Customized Image Generation}
Customized image generation has emerged as a pivotal task within the broader landscape of image synthesis, focusing on enabling pretrained diffusion models to understand specific concepts from limited data and re-render them in diverse contexts. Early optimization-based methods, such as Textual Inversion\cite{gal2022image} and DreamBooth\cite{ruiz2023dreambooth}, paved the way by learning specific tokens or fine-tuning the model for a single subject. 
Methods like ELITE\cite{wei2023elite}, IP-Adapter\cite{ye2023ip}, InstantID\cite{wang2024instantid}, and MoMA\cite{song2024moma} treat personalization as a vision-conditioned generation task by training specialized adapters. 
With the emergence of Diffusion Transformers (DiT)\cite{peebles2023scalable} like FLUX\cite{flux2024} and SD3\cite{esser2024scaling}, the paradigm has further evolved toward leveraging strong in-context capabilities. For instance, OmniControl\cite{xie2023omnicontrol} and EasyControl\cite{zhang2025easycontrol} adapt text-to-image models for precise personalized control, while unified models like Bagel\cite{deng2025emerging} attempt to harmonize understanding and generation. Furthermore, large-scale models such as FLUX Kontext\cite{labs2025flux}, Qwen-Image-Edit\cite{wu2025qwen}, and FLUX2\cite{flux-2-2025} also leverage in-context learning for image generation and editing. 
However, zero-shot adapters often fail on out-of-distribution effects, making custom LoRA training the reliable industrial standard. Yet, directly composing these LoRAs with acceleration modules (e.g., lightx2v\cite{lightx2v}) triggers severe feature interference and semantic drift. To resolve this, we distill multiple customized effects into a single, few-step unified LoRA, completely avoiding the conflicts inherent in multi-module composition.

\vspace{-0.3cm}
\subsection{Few-Step Generation}
To address the inference inefficiency of diffusion models, Consistency Models (CMs) \cite{song2023consistency} and their derivatives \cite{luo2023latent,wang2024phased,zhai2024motion} enable few-step generation by enforcing trajectory self-consistency. Recently, Distribution Matching Distillation (DMD)~\cite{dmd1,dmd2} established a superior paradigm by directly minimizing the divergence between the teacher and student distributions. Recent advances further elevate DMD: Decoupled-DMD~\cite{decoupledmd} enhances fine details via independent noise schedules, while DMDR~\cite{jiang2025distribution} and Flash-DMD\cite{chen2025flash} integrate reinforcement learning to incorporate external preference rewards safely, surpassing the teacher's performance ceiling.
Despite these advancements, existing DMD-based methods are largely confined to single-teacher, homogeneous settings. They suffer from severe training instability and feature collapse when bridging large student-teacher gaps or simultaneously fitting multiple target distributions. To overcome these bottlenecks, we propose CollectionLoRA, a multi-teacher distillation framework designed to stabilize multi-source matching and prevent distribution collapse.
\vspace{-0.3cm}
\subsection{On-Policy Distillation}
Standard offline distillation often suffers from exposure bias and compounding errors~\cite{llmopd}. OPD~\cite{llmopd,minillm,rethinkingopd} mitigates these issues by applying teacher feedback directly to states visited by the student’s own rollouts. While initially validated in Large Language Models for superior stability over scalar-reward reinforcement learning~\cite{rlopd}, OPD has recently been adapted to continuous visual generation. In diffusion and flow matching (e.g., Flow-OPD~\cite{fang2026flowopdonpolicydistillationflow}, D-OPSD~\cite{jiang2026dopsdonpolicyselfdistillationcontinuously}), OPD matches the teacher’s dense velocity fields along student-sampled trajectories.
Technically, while traditional OPD typically aligns the conditional transition distribution step-wise along trajectories, DMD~\cite{dmd2} focuses on optimizing the marginal data distribution of generated samples. Following recent literature~\cite{gu2026anyflowanystepvideodiffusion,chern2025livetalkrealtimemultimodalinteractive}, we conceptually unify DMD under the OPD taxonomy, as both frameworks fundamentally rely on correcting the student's on-policy explored states with teacher signals.

Building upon this paradigm, CollectionLoRA pioneers large-scale multi-teacher distillation to efficiently consolidate 50 to 180 diverse visual effects alongside few-step generation capabilities into a single, unified module.
\section{Preliminaries: Distribution Matching Distillation}
\label{sec:preliminaries}

Distribution Matching Distillation (DMD) aims to train an efficient student generator $G_\theta$ such that its generated distribution $p_{fake}$ approximates the target real distribution $p_{real}$ defined by a pre-trained diffusion teacher. To ensure consistency between training and inference in few-step synthesis, DMD2~\cite{dmd2} employs \textbf{Backward Simulation} to simulate the inference process. Specifically, starting from pure noise $z \sim \mathcal{N}(0, \mathbf{I})$, the simulation iteratively performs a sequence of denoising and re-noising steps: it predicts a denoised sample $\hat{x}_0$, then adds noise back to reach the next scheduled timestep. This iterative loop continues until a selected timestep $t$ is reached, yielding a simulated clean image that effectively captures the cumulative sampling characteristics of the multi-step inference trajectory. 

This simulated image then serves as the training target, replacing the conventional real data. The generator $G_\theta$ is trained to denoise a noisy version of this simulated sample to get the generated sample $x_g$, with the objective of matching the score functions of the real and fake distributions. The gradient for updating the generator parameters $\theta$ is formulated as:
\begin{equation}
    \label{eq: dmd}
    \nabla_\theta \mathcal{L}_{DMD} = \mathbb{E}_{z, t, \epsilon} \left[ \left( s_{fake}(x_t, t) - s_{real}(x_t, t) \right) \nabla_\theta x_g \right],
\end{equation}
where $x_t$ is the diffused state of the generated sample $x_g$. The target score $s_{real}$ is derived from the frozen teacher, while the fake score $s_{fake}$ is estimated by a critic model updated via standard denoising loss. 
\section{Method}
\label{sec:method}
To integrate dozens of heterogeneous visual effects and few-step generation capabilities into a single LoRA, we propose the CollectionLoRA framework, which aims to address parameter interference and deployment overheads via multi-teacher distillation. In Sec.~\ref{method:4_1}, we first formally define the general paradigm of visual effect LoRA training and analyze the challenges of multi-LoRA deployment. In Sec.~\ref{method:4_2}, we detail the Probabilistic Dual-Stream Routing mechanism, which leverages general-domain data as structural regularization to enhance model generalization in few-shot effect learning. To ensure the isolation of distinct concepts within a shared parameter space, we describe the Asymmetric Orthogonal Prompting  strategy in Sec.~\ref{method:4_3}. Finally, we present the Coarse-to-Fine Distillation Objective in Sec.~\ref{method:4_4} and the total training objective in Sec.~\ref{method:4_5}.

\begin{figure*}[!t]
    \centering
    \includegraphics[width=\linewidth]{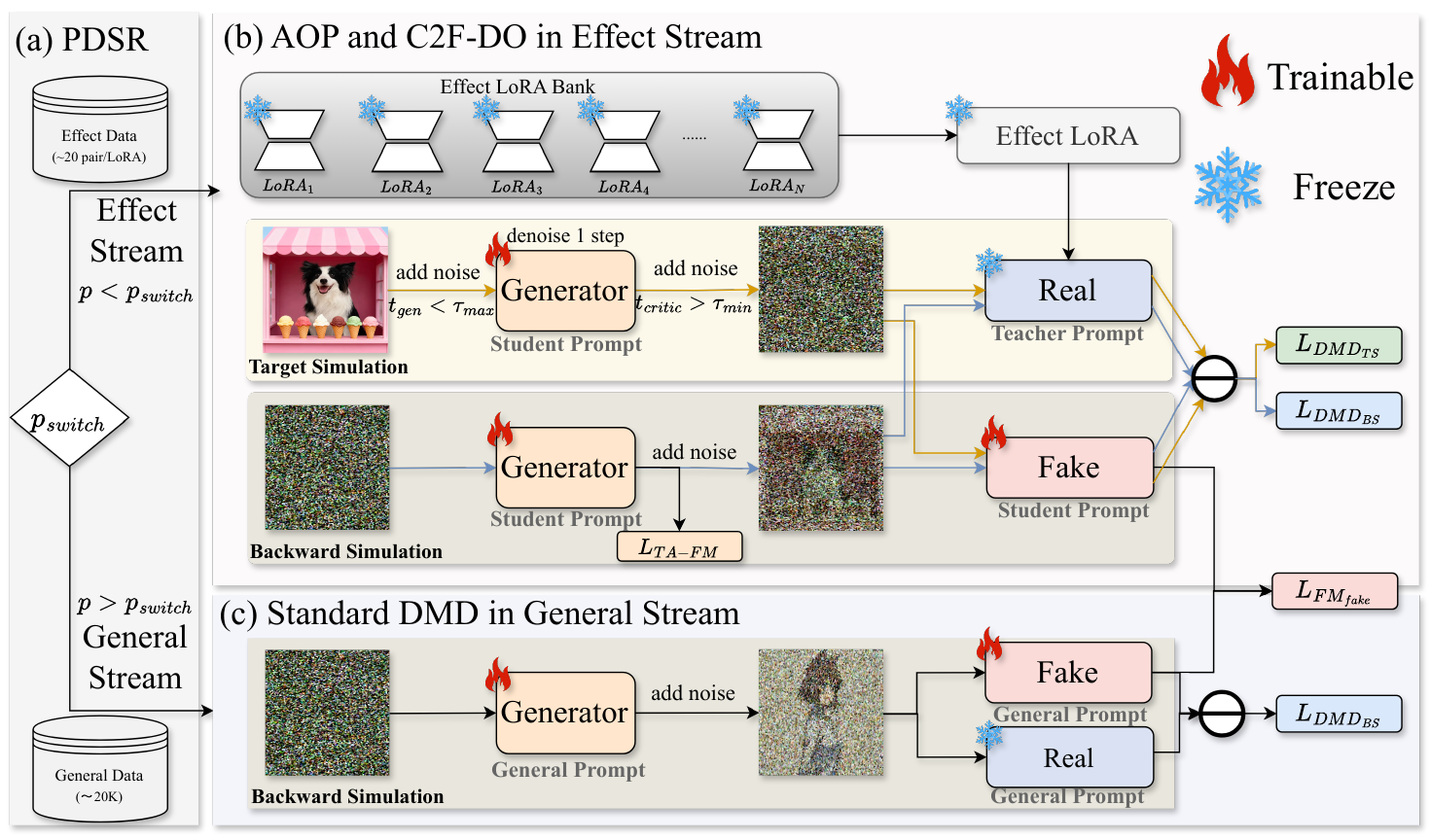}
    \caption{
    \textbf{The overall framework of CollectionLoRA.} (a) PDSR dynamically routes training batches into either the effect or general stream based on probability $p_{switch}$. (b) Effect Stream (with AOP \& C2F-DO): A frozen teacher LoRA is sampled from the bank. The trainable student generator uses asymmetric prompts and learns both target distributions and denoising trajectories via $\mathcal{L}_{\text{TA-FM}}$ for early-stage structural stabilization, alongside Target Simulation and Backward Simulation to restore fine details and match the overall distribution, while the trainable fake model is updated via $\mathcal{L}_{FM_{fake}}$.(c) General Stream: Performs standard DMD distillation on general data to prevent catastrophic forgetting and maintain foundational generalization.
    }
    \label{fig:framework}
\end{figure*}

\subsection{Problem Formulation}\label{method:4_1}
\paragraph{Standard Fine-Tuning for a Single Effect.}
For standard personalized fine-tuning of diffusion models, given the pre-trained base model parameters $\theta_{base}$ and a limited set of paired data for a specific effect $effect_i$, Low-Rank Adaptation (LoRA)~\cite{LoRA} is typically employed to learn the effect-specific residual weights $\Delta\theta_{effect}^{i}$.
The training is generally optimized via the Flow Matching loss~\cite{lipman2023flowmatchinggenerativemodeling} by regressing the target vector field:
\begin{equation}
    \mathcal{L}_{FM} = \mathbb{E}_{x_0, c, \epsilon, t} [||v_{\theta_{base} + \Delta\theta_{effect}^{i}}(x_t, t, c) - (x_0 - \epsilon)||_2^2],
\end{equation}
where $x_0$ represents the ground-truth target effect image, $\epsilon \sim \mathcal{N}(0, I)$ is the sampled standard Gaussian noise, $t \in [0, 1]$ denotes the continuous time step, $c$ represents the conditioning input, comprising the editing instruction and the source reference image.
\paragraph{Dilemma of Multi-Module Composition.}
In practical inference, achieving efficient few-step sampling necessitates retrieval and composition based on user instructions as shown in Fig.~\ref{fig:intro}(a). The deployment model weights $\theta_{deploy}$ are formulated as:
\begin{equation}
    \theta_{deploy}=\theta_{base}+\text{Retrieve}(\mathcal{B},\text{instruction})+\Delta\theta_{acc},
\end{equation}
where $\text{Retrieve}(\cdot)$ denotes the process of retrieving the corresponding effect LoRA from the effect LoRA bank $\mathcal{B}$ by the editing instruction, which significantly increases routing latency and the risk of matching errors as the bank scales. More crucially, interactions between the acceleration LoRA $\Delta\theta_{acc}$ and the retrieved effect LoRA lead to severe concept bleeding, semantic drift, and degradation in style fidelity.
\paragraph{Proposed Paradigm: CollectionLoRA.}
To overcome the aforementioned limitations in deployment and composition, we propose universally distilling $N$ heterogeneous visual effects, along with few-step acceleration capabilities, into a single student LoRA $\Delta\theta_{student}$.
In our multi-teacher distillation framework, we instantiate a set of effect teachers $\mathcal{T}=\{T_{effect}^{1},T_{effect}^{2},\dots,T_{effect}^{N}\}$ by equipping the base model with various single-effect LoRAs.
The objective of the student generator $G_{\theta}$ is to fit the high-quality target distributions $y$ generated by all teacher models.
During final deployment, the inference process is significantly simplified, eliminating the need to dynamically load and compose multiple LoRAs:
\begin{equation}
    x_{g}=G_{\theta}(\epsilon,c_{student}),\quad\text{where}\quad\theta=\theta_{base}+\Delta\theta_{student}.
\end{equation}
This paradigm not only eliminates runtime routing overhead but also fundamentally resolves compositional conflicts among modules at the parameter level as shown in Fig.~\ref{fig:intro}(b).

\subsection{Probabilistic Dual-Stream Routing}\label{method:4_2}
To enhance the robustness and generalization of the model in effect generation tasks, we design a Probabilistic Dual-Stream Routing (PDSR) mechanism, as shown in Fig.~\ref{fig:framework}(a).
Specifically, the framework samples a random probability $p \sim \mathcal{U}(0,1)$ at each training step and dynamically executes the following routing logic based on a preset switching rate $p_{switch}$:
\paragraph{General Stream ($p \ge p_{switch}$):} This stream utilizes unlabeled general-domain images, employing the frozen base model $\theta_{base}$ as the teacher. The distribution matching loss $\mathcal{L}_{\text{DMD\_BS}}$ is calculated via a standard backward simulation mechanism in Eq.~\ref{eq: dmd}.
\paragraph{Effect Stream ($p < p_{switch}$):}
This stream focuses on the precise injection of the $N$ effect capabilities. The system dynamically loads the effect LoRA $\Delta\theta_{effect}^{i}$ to instantiate a specific effect teacher $T_{effect}^{i}$. 
We leverage a Coarse-to-Fine Distillation Objective (C2F-DO) to address the significant early-stage distribution discrepancies between the teacher and student models during heterogeneous distillation. The specific mechanisms will be detailed in Section \ref{method:4_4}.

\subsection{Asymmetric Orthogonal Prompting }\label{method:4_3}
To mitigate feature interference and concept leakage during multi-effect integration, we propose the Asymmetric Orthogonal Prompting  (AOP) strategy. Unlike standard distillation, AOP uses different prompts for the teacher and student models to ensure clear concept isolation. Specifically, each effect teacher $T_{\text{effect}}^{i}$ uses its original training prompt $c_{\text{teacher}}^{i}$ to generate high-quality target images $y$. To avoid semantic confusion in the student model, we use a Vision-Language Model (VLM) to automatically generate a descriptive caption $c_{\text{vlm}}^{i}$ for each effect. This automated process removes the need for manual prompt engineering. We then assign a unique orthogonal trigger word $v^{i}$ to each effect and construct the student condition as:
\begin{equation}
c_{\text{student}}^{i} = [v^{i}, c_{\text{vlm}}^{i}].
\end{equation}

\begin{figure*}[!t]
    \centering
    \includegraphics[width=\linewidth]{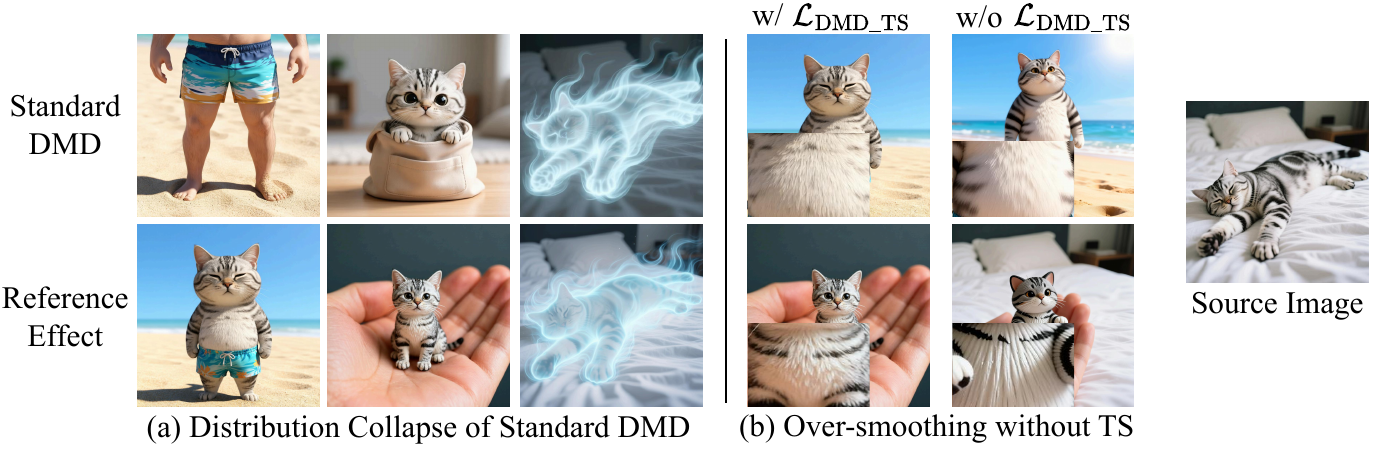}
    \caption{\textbf{Effectiveness of C2F-DO.} (a) Directly applying standard DMD to multi-teacher distillation causes the student distribution to collapse into an intermediate state. (b) Relying solely on trajectory anchoring leads to detail loss in high-frequency features, whereas incorporating target simulation effectively restores realistic microscopic details.}
    \label{fig:method_4_4}
\end{figure*}

\subsection{Coarse-to-Fine Distillation Objective}\label{method:4_4}
In the effect stream, due to the substantial distribution discrepancy between the student and teacher models during the early training stages, the vanilla DMD objective causes the student distribution to collapse into an intermediate manifold between the original and effect distributions, resulting in training failure, as illustrated in Fig.~\ref{fig:method_4_4}(a).
To address this, we propose the Coarse-to-Fine Distillation Objective (C2F-DO), which synergizes trajectory anchoring and distribution matching to stabilize the optimization process and ensure generation quality.
\paragraph{Trajectory Anchoring via Flow Matching (TA-FM).}
To bridge the initial distribution gap, we formulate a flow matching objective, guiding the student model to optimize along the trajectory pointing towards the target image $y$:
\begin{equation}
\mathcal{L}_{\text{TA-FM}} = \left| G_{\theta}(y_{t}, t, c_{\text{student}}) - (y - \epsilon) \right|_2^2,
\end{equation}
where $y_{t}=ty+(1-t)\epsilon$ denotes the linear interpolation state.
While $\mathcal{L}_{\text{TA-FM}}$ provides a stable optimization direction in the early training stages, we observe that over-reliance on it leads to excessive smoothing of high-frequency image textures due to its regression nature, as illustrated in Fig.~\ref{fig:method_4_4}(b).
\paragraph{Target-Simulated Distribution Matching.}
We introduce Target Simulation (TS) on the target image $y$. By aligning student and teacher score functions, this divergence-minimization objective compels the model to capture statistical variance, effectively restoring high-frequency features.
In this branch, $y$ is diffused to time step $t_{\text{gen}}$, denoised by the generator to obtain the simulated output $\hat{y}$, and subsequently re-noised to $t_{\text{critic}}$.
The update gradient is formulated as:
\begin{equation}
\nabla_{\theta}\mathcal{L}_{\text{DMD\_TS}} = \mathbb{E}_{t_{\text{gen}} < \tau_{\text{max}}, t_{\text{critic}} > \tau_{\text{min}}, \epsilon} \left[ \left( s_{\text{fake}}(\hat{y}_{t_{\text{critic}}}, t_{\text{critic}}) - s_{\text{real}}(\hat{y}_{t_{\text{critic}}}, t_{\text{critic}}) \right) \nabla_{\theta}\hat{y} \right].
\end{equation}
We impose strict dual-timestep sampling constraints:
\begin{itemize}
    \item \textbf{Generator Upper Bound ($t_{\text{gen}} < \tau_{\text{max}}$).} Restricts the forward diffusion depth to preserve the prior of the effect teacher, preventing the denoising process from deviating excessively and degrading into unguided free generation.
    \item \textbf{Critic Lower Bound ($t_{\text{critic}} > \tau_{\text{min}}$).} Ensures that sufficient noise $\epsilon$ is injected into $\hat{y}$ during the evaluation phase, thereby fully amplifying the divergence between the real score $s_{\text{real}}$ and the fake score $s_{\text{fake}}$ to provide reliable gradient guidance for $\Delta\theta_{\text{student}}$.
\end{itemize}
The comprehensive optimization objective for the effect stream is:
\begin{equation}
\mathcal{L}_{\text{C2F-DO}} = \mathcal{L}_{\text{TA-FM}} + \mathcal{L}_{\text{DMD\_TS}} + \mathcal{L}_{\text{DMD\_BS}}.
\end{equation}
Concurrently, $\mathcal{L}_{\text{DMD\_BS}}$ acts as a persistent regularizer, ensuring the student retains the teacher's global style distribution.

\subsection{Overall Objective}\label{method:4_5}
Driven by the dynamic routing of the PDSR mechanism at each iteration, the final overall optimization objective $\mathcal{L}_{\text{total}}$ is formulated as:
\begin{equation}
\mathcal{L}_{\text{total}} = \mathbb{1}_{\{\text{general}\}} \mathcal{L}_{\text{DMD\_BS}} + \mathbb{1}_{\{\text{effect}\}} \mathcal{L}_{\text{C2F-DO}},
\end{equation}
where the indicator functions $\mathbb{1}_{\{\text{general}\}}$ and $\mathbb{1}_{\{\text{effect}\}}$ are mutually exclusive, strictly determined by the routing state at the current step.
When routed to the general stream, the model computes only $\mathcal{L}_{\text{DMD\_BS}}$ via backward simulation to consolidate fundamental priors, and when routed to the effect concept stream, it applies $\mathcal{L}_{\text{C2F-DO}}$ for the injection of effect capabilities. Throughout this process, the model naturally acquires few-step generation capabilities.

\section{Experiment}
\label{sec:exp}

\vspace{-0.1cm}
\subsection{Experimental Setup}
\label{subsec:experimental_setup}
\textbf{Datasets.} 
Our framework utilizes two datasets for training: an effect dataset comprising 50 specific effects (each with ~20 animal/portrait image pairs), and a general dataset of 20K source images paired with MLLM-generated instructions, requiring no target images.
For evaluation, we introduce EffectBench. Aligned with our training data, it comprises animal and portrait categories. We use Gemini-2.5 Pro and Qwen-Image \cite{qwenimage} to generate 100 diverse test images per category, ensuring high variance in subject types, actions, scenes, and camera distances. This yields an evaluation protocol of 5,000 instructions per model.

\noindent
\textbf{Baseline Methods.} We adopt Qwen-Image-Edit-2509 \cite{qwenimage} as the base model and compare our approach against two standard paradigms: (1) Base Model + Effect LoRA, and (2) Base Model + Effect LoRA + Acceleration LoRA. For acceleration, we utilize the popular Qwen-Image-Edit-Lightning LoRA released by lightx2v\cite{lightx2v}. To evaluate multi-concept injection capabilities, we also construct a strong baseline denoted as 50-in-1 (FM), which optimizes a unified LoRA on all aggregated training data using a standard flow matching objective.

\begin{figure*}[!t]
    \vspace{-0.4cm}
    \centering
    \includegraphics[width=\linewidth]{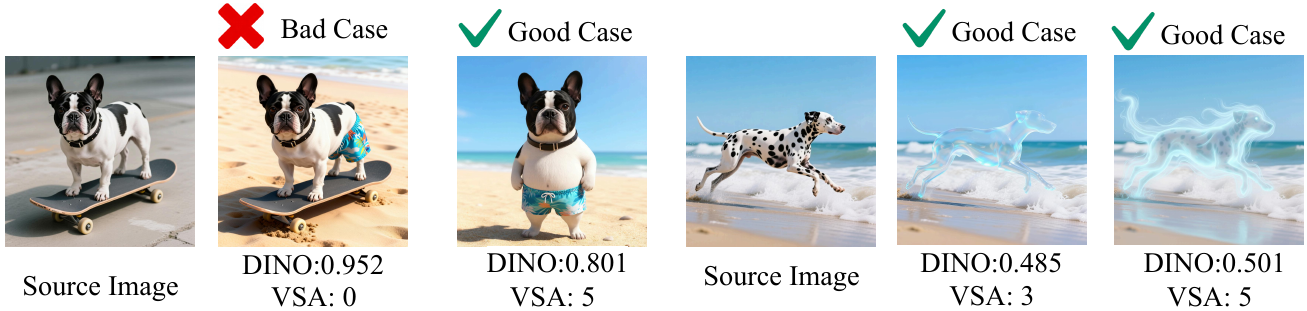}
    \caption{\textbf{Evaluation of subject consistency metrics.} While DINO often assigns high scores to failed generations based on superficial similarity, VSA strictly penalizes semantic failures with a zero score, providing a robust measure of actual subject consistency in complex stylizations.}
    \label{fig:metric}
    \vspace{-0.4cm}
\end{figure*}

\noindent
\textbf{Evaluation Metrics.} We assess generation quality using several metrics: CLIP \cite{clip} and DreamSim \cite{fu2023dreamsimlearningnewdimensions} for style alignment, DINO \cite{zhang2022dinodetrimproveddenoising} for subject consistency, and EditReward \cite{wu2026editrewardhumanalignedrewardmodel} for instruction-following and overall image quality. We also use an MLLM to calculate the Bad Case Rate (BCR), representing the proportion of effect failures across all test images.
Because traditional metrics like DINO often fail to capture subject consistency in complex effect scenarios (Fig.~\ref{fig:metric}), we introduce the Valid Subject Alignment (VSA) metric. VSA uses the MLLM for a two-stage evaluation: it first verifies if the target effect was successfully applied. If it fails, the consistency score defaults to zero and if successful, the MLLM evaluates the underlying subject consistency. Further details are provided in the appendix.

\noindent
\textbf{Implementation Details.} All effect LoRA training is performed at a 1024x1024 resolution using the ai-toolkit \cite{aitookit} with a learning rate of 1e-4. Single-effect LoRAs are trained for 2,000 steps, while the 50-in-1 (FM) baseline is trained for 30,000 steps.
In our method, both the generator and the fake score model are fine-tuned exclusively via LoRA with a learning rate of 1e-4. The update frequency ratio of the fake score model to the generator is 5:1, and the generator is trained for 5,000 steps. By default, the stream switching probability $p_{\text{switch}}$ is 0.5, with timestep bounds $\tau_{\text{max}} = 750$ and $\tau_{\text{min}} = 500$. All experiments are conducted on 8 NVIDIA H800 GPUs.

\vspace{-0.3cm}
\subsection{Quantitative Evaluation}
\label{subsec:quantitative}

\begin{table}[!t]
\centering
\caption{Quantitative comparison on our EffectBench.}
\label{tab:main_results}
\resizebox{\columnwidth}{!}{ 
\begin{tabular}{ccccccccc} 
\toprule
\textbf{Setting} & \textbf{Method} & \textbf{CLIP} ($\uparrow$) & \textbf{DreamSim} ($\downarrow$) & \textbf{DINO} ($\uparrow$) & \textbf{VSA} ($\uparrow$) & \textbf{EditReward} ($\uparrow$) & \textbf{BCR} ($\downarrow$) & \textbf{NFE} ($\downarrow$) \\
\midrule
\multirow{2}{*}[-2pt]{\makecell{Single \\ Effect}} & \makecell{Base} & 0.726 & 0.434 & 0.611 & 4.075 & 1.007 & 0.141 & 40 $\times$  2 \\ 
\cmidrule(lr){2-9} 
 & \makecell{Base+Lightning} & 0.717 & 0.441 & \textbf{0.612} & 3.901 & 0.986 & 0.168 & 8 \\ 
\midrule
\multirow{2}{*}[-2pt]{\makecell{50 Effects \\ in 1}} & \makecell{FM + Lightning} & 0.703 & 0.468 & 0.611 & 4.150 & 0.929 & 0.217 & 8 \\ 
\cmidrule(lr){2-9} 
 & \makecell{\textbf{Ours}} & \textbf{0.727} & \textbf{0.425} & 0.600 & \textbf{4.380} & \textbf{1.052} & \textbf{0.087} & \textbf{8} \\ 
\bottomrule
\end{tabular}
}
\end{table}

\begin{table}[!t]
\centering
\caption{Deployment costs across numbers of LoRAs.}
\label{tab:deploy_cost}
\resizebox{\columnwidth}{!}{%
\begin{tabular}{ccccccc}
\toprule
Metric & Method & 10 LoRAs & 20 LoRAs & 50 LoRAs & 100 LoRAs & 150 LoRAs \\ 
\midrule
\multirow{2}{*}{Routing Latency} 
& baseline & 6.88s/q & 6.95 s/q & 7.09s/q & 7.22s/q & 9.18s/q \\ \cmidrule{2-7} 
& \textbf{ours} & \textbf{0s/q} & \textbf{0s/q} & \textbf{0s/q} & \textbf{7.22s/q} & \textbf{9.18s/q} \\ 
\midrule
\multirow{2}{*}{\makecell{LoRA Loading Latency \\ $\times$ Switch Count}} 
& baseline & 1.2s*200 & 1.2s*200 & 1.2s*200 & 1.2s*200 & 1.2s*200 \\ \cmidrule{2-7} 
& \textbf{ours} & \textbf{0s} & \textbf{0s} & \textbf{0s} & \textbf{1.2s*108} & \textbf{1.2s*136} \\ 
\midrule
\multirow{2}{*}{Routing Accuracy} 
& baseline & 99\% & 94\% & 87\% & 85\% & 76\% \\ \cmidrule{2-7} 
& \textbf{ours} & \textbf{100\%} & \textbf{100\%} & \textbf{100\%} & \textbf{90\%} & \textbf{82\%} \\ 
\midrule
\multirow{2}{*}{Storage Overhead} 
& baseline & 2.2G * 10 & 2.2G * 20 & 2.2G * 50 & 2.2G * 100 & 2.2G * 150 \\ \cmidrule{2-7} 
& \textbf{ours} & \textbf{2.2G} & \textbf{2.2G} & \textbf{2.2G} & \textbf{2.2G * 2} & \textbf{2.2G * 3} \\ 
\bottomrule
\end{tabular}%
}
\vspace{-0.3cm}
\end{table}

\textbf{Quantitative Comparison with Baselines.}
Table \ref{tab:main_results} compares our joint distillation approach (50 Effects in 1, NFE=8) against independent single-effect models (NFE=80) and a naive multi-task baseline (FM + Lightning). Our method achieves state-of-the-art style alignment and overall quality (CLIP: 0.727, DreamSim: 0.425, EditReward: 1.052), remarkably outperforming the computationally expensive single-task teacher and mitigating the degradation typical of multi-concept fusion. Benefiting from our AOP and C2F-DO strategies, we effectively suppress parameter interference, yielding a Bad Case Rate (BCR) of just 0.087—substantially lower than both the baseline (0.217) and the teacher (0.141). Regarding subject consistency, although a marginal decrease is observed in the traditional DINO metric (0.600 vs. 0.611)—which often struggles with severe visual transformations as shown in Fig.~\ref{fig:metric}—our method achieves the highest Valid Subject Alignment (VSA) score (4.380). This demonstrates that our unified model robustly triggers effects, minimizes failure penalties, and better preserves foundational structures under extreme concept compression.

\noindent
\textbf{Deployment Cost Analysis.}
To evaluate deployment overhead, we simulated 200 queries on a single GPU against a VLM-routed baseline. For 10-50 LoRAs, CollectionLoRA bypasses routing entirely, achieving 0s latency, 100\% accuracy, and a constant 2.2GB storage overhead, circumventing the baseline's linear storage growth and accuracy degradation. At 100-150 LoRAs, our approach reverts to VLM routing but still drastically reduces storage to 2\% of the baseline, decreases model switches (136 vs. 200), and maintains higher accuracy (82\% vs. 76\%). Consequently, its minimal memory footprint and switching burdens highlight significant cost-saving potential for large-scale multi-GPU deployments.

\begin{figure*}[!t]
    \centering
    \includegraphics[width=\linewidth]{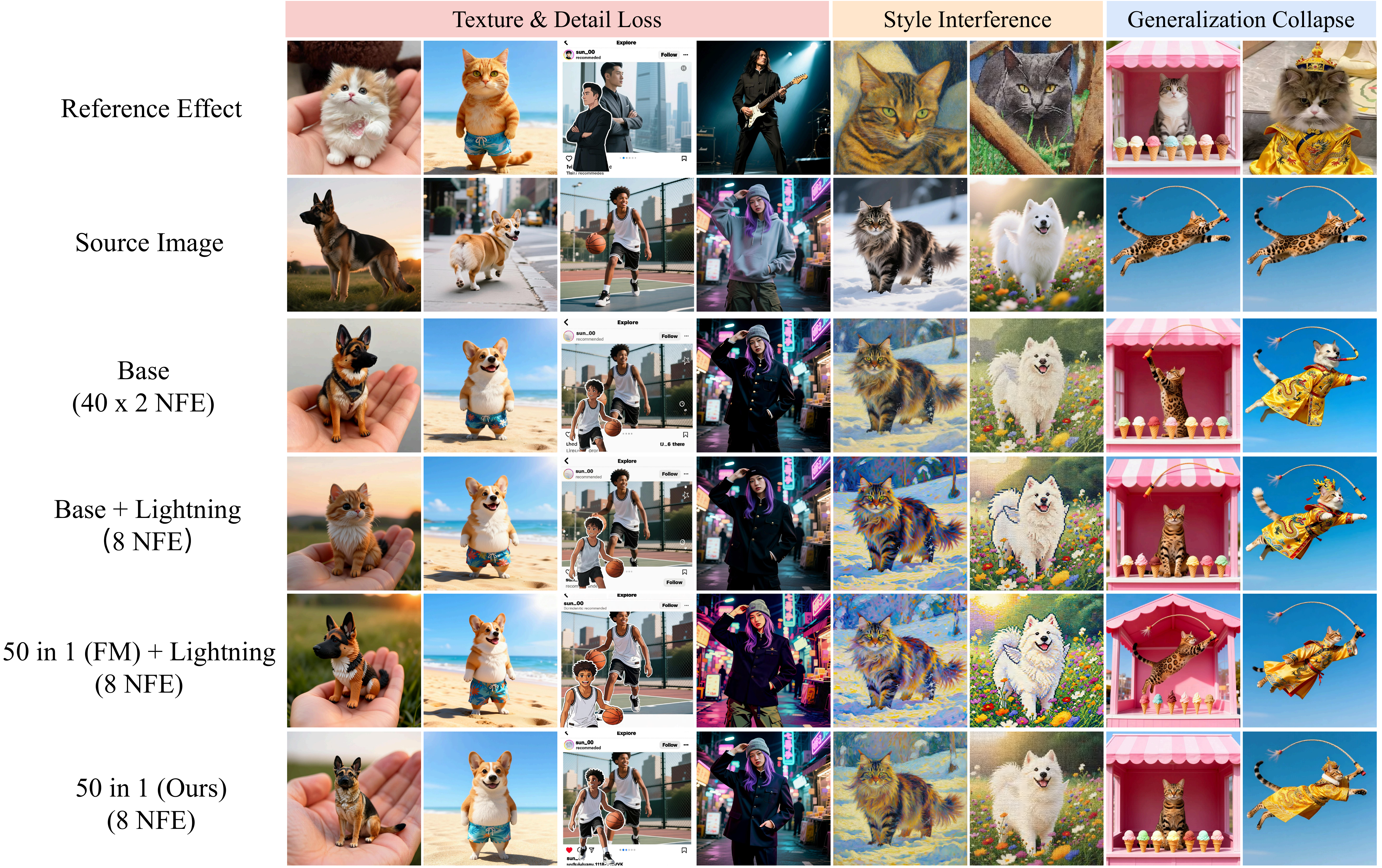}
    \caption{\textbf{Qualitative comparison of CollectionLoRA against baseline methods.} The visual results indicate that CollectionLoRA effectively mitigates texture and detail loss, style interference, and generalization collapse commonly observed in the alternative pipelines.}
    \vspace{-0.5cm}
    \label{fig:main_compare}
\end{figure*}

\vspace{-0.2cm}
\subsection{Qualitative Evaluation}
\label{subsec:qualitative}

\textbf{Qualitative Comparison.} Fig~\ref{fig:main_compare} compares qualitative results. In the \textbf{Texture \& Detail Loss panel}, the accelerated baseline (Base+Lightning) exhibits semantic drift (e.g., identity shifts, col. 1) and structural degradation (e.g., detail loss, col. 3). Furthermore, the joint distillation baseline (FM+Lightning) oversmooths due to regression losses, losing high-frequency textures (e.g., pet fur, cols. 1-2) and realism. In contrast, our constrained Target Simulation mechanism restores fine-grained textures and physical realism while preserving structural consistency.
The \textbf{Style Interference panel} shows style degradation from module serialization and parameter crowding. For Base+Lightning, directly concatenating acceleration and effect LoRAs disrupts the feature manifold and degrades style purity. Second, without concept isolation, FM+Lightning outputs (cols. 5-6) show concept bleeding from the color palettes and brushstrokes of other effects. 
CollectionLoRA resolves this by isolating latent effects, synthesizing pure, crosstalk-free styles even under 50-effect, 8-step inference.
The \textbf{Generalization Collapse panel} illustrates the generalization limits of few-shot fine-tuning. Due to minimal training data, baseline models yield distorted structures and poses for out-of-distribution (OOD) inputs (e.g., the flying cat). By using a dual-stream routing mechanism to introduce general-domain data as structural regularization, our method preserves the base model's generalization and structural fidelity for rare subjects.

\begin{figure*}[!t]
    \centering
    \includegraphics[width=\linewidth]{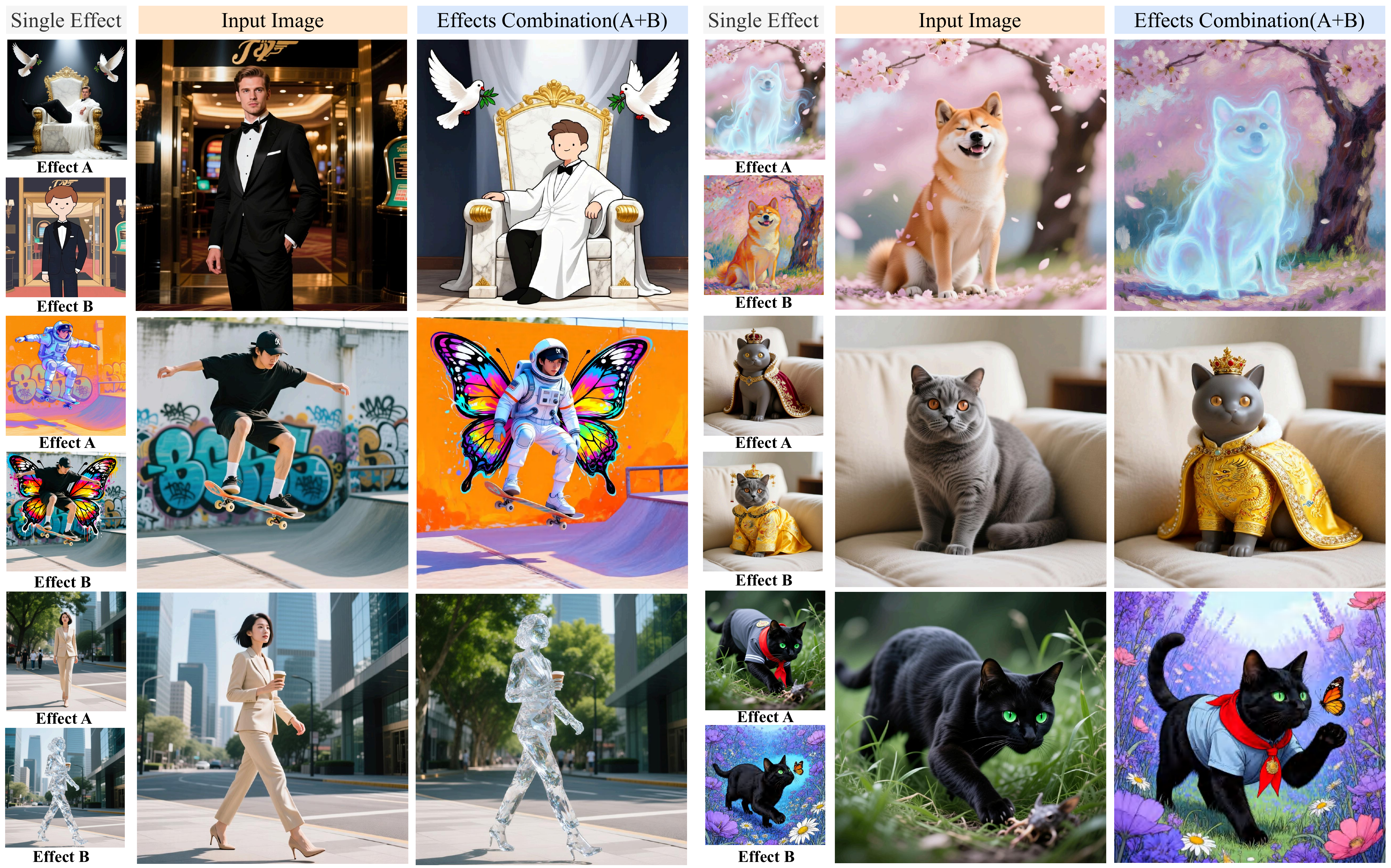}
    \caption{\textbf{Zero-shot effect composition capability of CollectionLoRA.} Given two independently learned effects, CollectionLoRA can simultaneously apply both to an input image via a compositional prompt (\emph{``Please apply \{Effect A\} to the input image, and then apply \{Effect B\}.''}), without any additional training.}
    \vspace{-0.7cm}
    \label{fig:AB_test}
\end{figure*}

\noindent
\textbf{Zero-Shot Inference of Effects Combinations.} 
Beyond reproducing individual effects, we discover an emergent compositional capability in CollectionLoRA that is never explicitly trained for. By simply chaining two effect descriptors in a single instruction, our model can simultaneously activate and blend two distinct effects in one forward pass \textbf{without any additional fine-tuning}.
As shown in Fig.~\ref{fig:AB_test}, the composed outputs faithfully inherit the visual characteristics of both effects while preserving subject identity. We attribute this behavior to our AOP strategy, which encodes each effect into an orthogonal subspace of the prompt manifold, enabling disentangled representations that can be compositionally activated at inference time. This finding suggests that CollectionLoRA constructs a structured effect space where concepts are independently controllable and combinable, substantially expanding its expressive capacity beyond the explicitly trained effect vocabulary.

\subsection{Ablation Study}
\label{sec:ablation}


\begin{table*}[t]
    \centering
    \caption{Ablation study of the proposed components under the 50-in-1 concurrent setting. \checkmark indicates the application of the specific module. The best results for each metric are highlighted in \textbf{bold}.}
    \label{tab:ablation}
    \resizebox{\textwidth}{!}{ 
    \begin{tabular}{ccccccccccc}
        \toprule
        Exp. & \makecell{PDSR} & \makecell{AOP}& \makecell{TS} & \makecell{TA-FM} & CLIP $\uparrow$ & DreamSim $\downarrow$ & DINO $\uparrow$ & VSA $\uparrow$ & EditReward $\uparrow$ & BCR $\downarrow$ \\
        \midrule
        (1) & \checkmark & & & & 0.725 & 0.434 & 0.514 & 2.756 & 0.989 & 0.378 \\
        (2) & \checkmark & \checkmark & & & 0.732 & 0.427 & 0.525 & 3.720 & 1.008 & 0.207 \\
        (3) & \checkmark & \checkmark & \checkmark & & \textbf{0.736} & \textbf{0.420} & 0.541 & 4.018 & 0.979 & 0.199 \\
        (4) & & \checkmark & \checkmark & \checkmark & 0.727 & 0.426 & 0.590 & 4.248 & 0.976 & 0.108 \\
        (5) & \checkmark & \checkmark & \checkmark & \checkmark & 0.727 & 0.425 & \textbf{0.600} & \textbf{4.380} & \textbf{1.052} & \textbf{0.087} \\
        \bottomrule
    \end{tabular}
    }
\end{table*}
\begin{figure*}[!t]
    \centering
    \includegraphics[width=\linewidth]{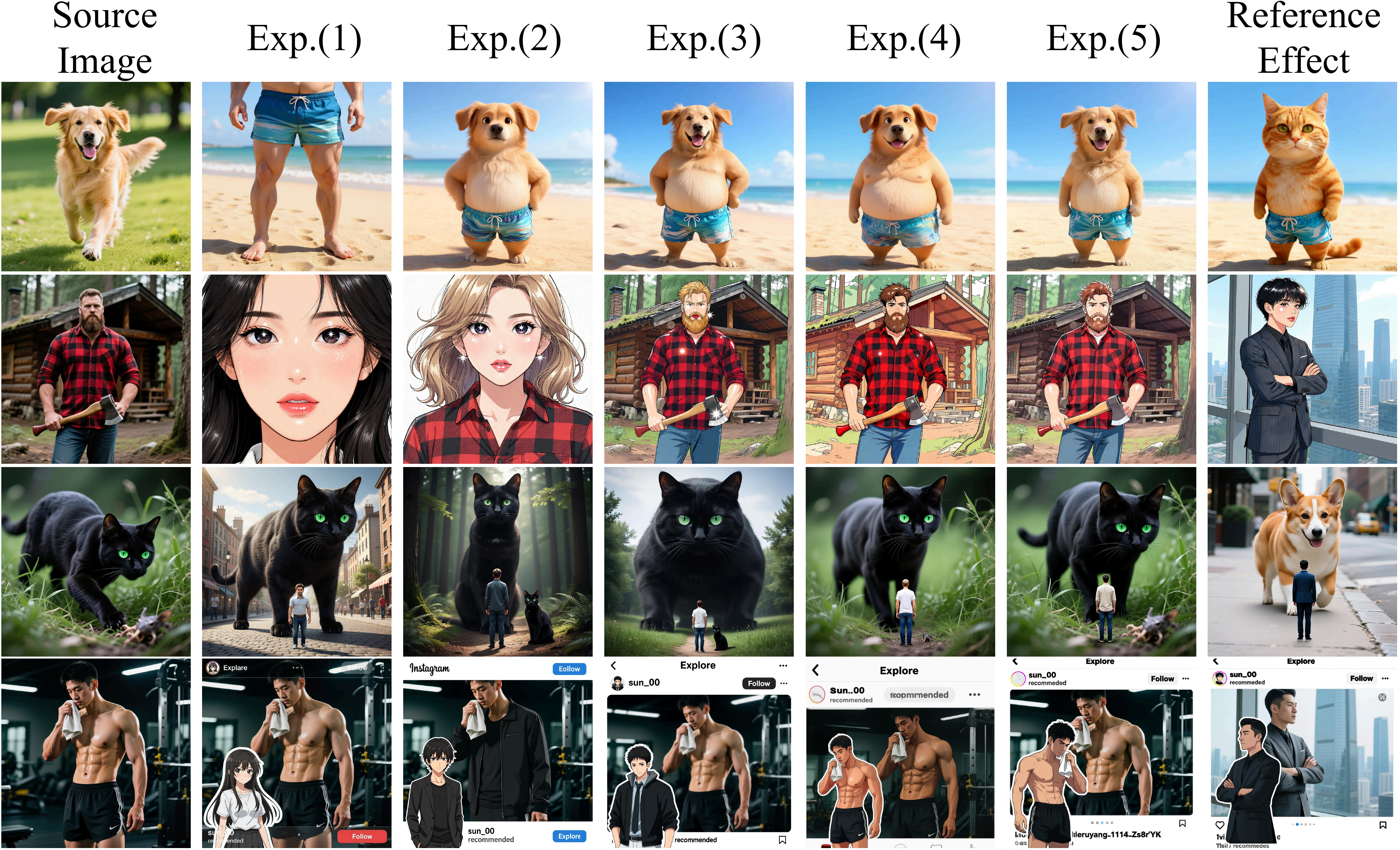}
    \captionof{figure}{\textbf{Qualitative ablation study.} The progressive integration of our core components systematically resolves semantic collapse, restores high-frequency textures, and ensures strict structural consistency.}
    \label{fig:ablation_comp}
    \vspace{-0.4cm}
\end{figure*}
\textbf{Quantitative Performance Analysis.} Under a 50-in-1 concurrent distillation setting (Tab.~\ref{tab:ablation}), our ablation validates each component: \textbf{AOP} mitigates concept bleeding, reducing BCR from 0.378 to 0.207; \textbf{TS} overcomes oversmoothing bias, achieving top CLIP (0.736) and DreamSim (0.420) scores; \textbf{TA-FM} stabilizes optimization, boosting VSA to 4.380 and minimizing BCR to 0.087; and \textbf{PDSR} prevents catastrophic forgetting, restoring EditReward to 1.052.

\noindent
\textbf{Qualitative Assessment}
Fig.~\ref{fig:ablation_comp} presents visual comparisons of our method. Exp.~(1) lacks AOP and suffers from semantic collapse. AOP decouples the effect and general streams and this separation ensures accurate task triggering shown in Exp.~(2). TS restores high-frequency textures in Exp.~(3). These details include skin and garment folds. Standard flow matching typically blurs such textures. TA-FM significantly improves structural consistency in Exp.~(5). It provides stable microscopic anchoring. Consequently, the model preserves the spatial layout and avoids pose distortions. Finally, PDSR resolves the background blending issues in Exp.~(4). This module achieves higher visual harmony between effects and complex environments.

\begin{table}[!t]
\centering
\caption{\textbf{CLIP Score} comparison across numbers of LoRAs}
\label{tab:lora_scale}
\resizebox{\columnwidth}{!}{%
\begin{tabular}{lccccc}
\toprule
 & \textbf{10 LoRAs} & \textbf{20 LoRAs} & \textbf{50 LoRAs} & \textbf{100 LoRAs} & \textbf{180 LoRAs} \\ 
\midrule
Base & 0.735 & 0.724 & 0.726 & \textbf{0.723} & \textbf{0.724} \\ 
Base+Lightning & 0.716 & 0.712 & 0.717 & 0.717 & 0.722 \\ 
All in 1 (FM) + Lightning & 0.725 & 0.722 & 0.703 & 0.694 & 0.689 \\ 
All in 1 (Ours) & \textbf{0.741} & \textbf{0.723} & \textbf{0.727} & 0.716 & 0.709 \\ 
\bottomrule
\end{tabular}%
}

\end{table}

\begin{table}[!t]
\centering
\caption{\textbf{CLIP score} comparison for incremental effect addition}
\vspace{-0.3cm}
\label{tab:lora_50_to_54}
\begin{tabular}{lccccc}
\toprule
 & 50 LoRAs & 51 LoRAs & 52 LoRAs & 53 LoRAs & 54 LoRAs \\ 
\midrule
Base+Lightning & 0.717 & 0.720 & 0.721 & 0.724 & 0.724 \\ 
\textbf{Ours} & \textbf{0.727} & \textbf{0.726} & \textbf{0.728} & \textbf{0.727} & \textbf{0.725} \\ 
\bottomrule
\end{tabular}%
\vspace{-0.4cm}
\end{table}

\noindent
\textbf{Lora Scaling \& Upper Bound.}
Tab.~\ref{tab:lora_scale} reports CLIP scores as the number of effects scales from 10 to 180. Our method consistently outperforms All-in-1 (FM) + Lightning across all settings, demonstrating that our hybrid distillation objective effectively mitigates the quality degradation caused by naive multi-concept fusion. At smaller scales (10--50 effects), our method even surpasses all baselines including the single-task Base model. As the number of effects increases to 100--180, a moderate performance drop is observed, which is expected given the increased optimization difficulty under extreme concept compression. Nevertheless, our method maintains competitive performance against Base+Lightning, confirming that CollectionLoRA scales gracefully to a large number of effects without catastrophic quality degradation.

\noindent
\textbf{Incremental Effect Extension.}
Tab.~\ref{tab:lora_50_to_54} validates the incremental extension capability of CollectionLoRA. Starting from the 50-effect model, adding effects $51_{st}$-$54_{th}$ via lightweight fine-tuning (100 steps for generator) consistently outperforms Base+Lightning across all settings, with CLIP scores remaining stable in the range of 0.725-0.728. The absence of catastrophic forgetting confirms that CollectionLoRA supports effect extension without retraining from scratch.

\begin{figure*}[!t]
    \centering
    \includegraphics[width=\linewidth]{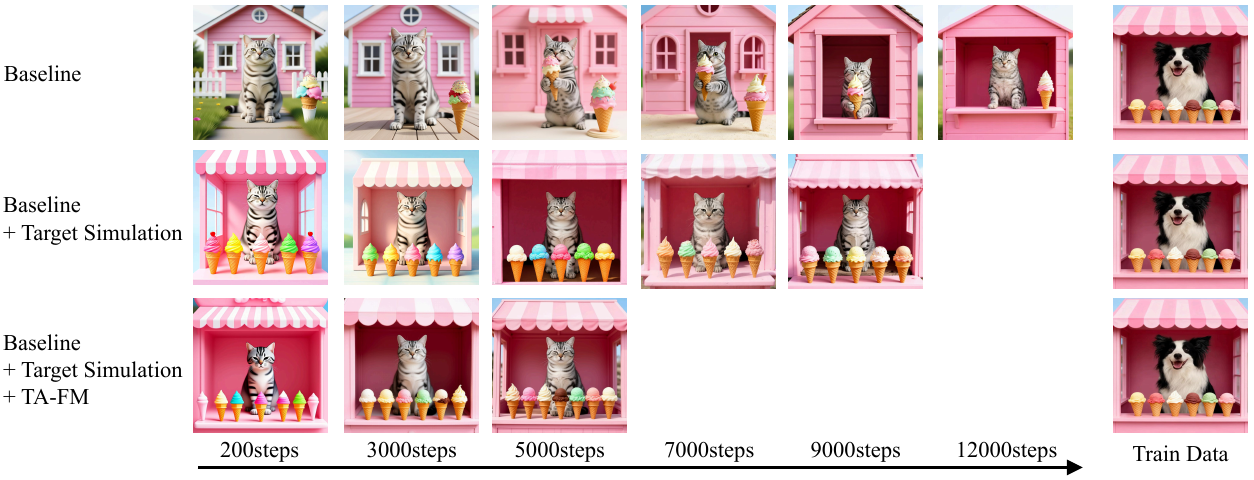}
    \captionof{figure}{\textbf{Qualitative ablation of training dynamics.} Integrating TA-FM and TS significantly accelerates structural convergence and enhances generative fidelity compared to the baseline.}
    \label{fig:ablation_trainsteps}
    \vspace{-0.1cm}
\end{figure*}
\begin{figure}[htbp]
    \centering
    \begin{subfigure}{0.48\linewidth}
        \centering
        \includegraphics[width=\linewidth]{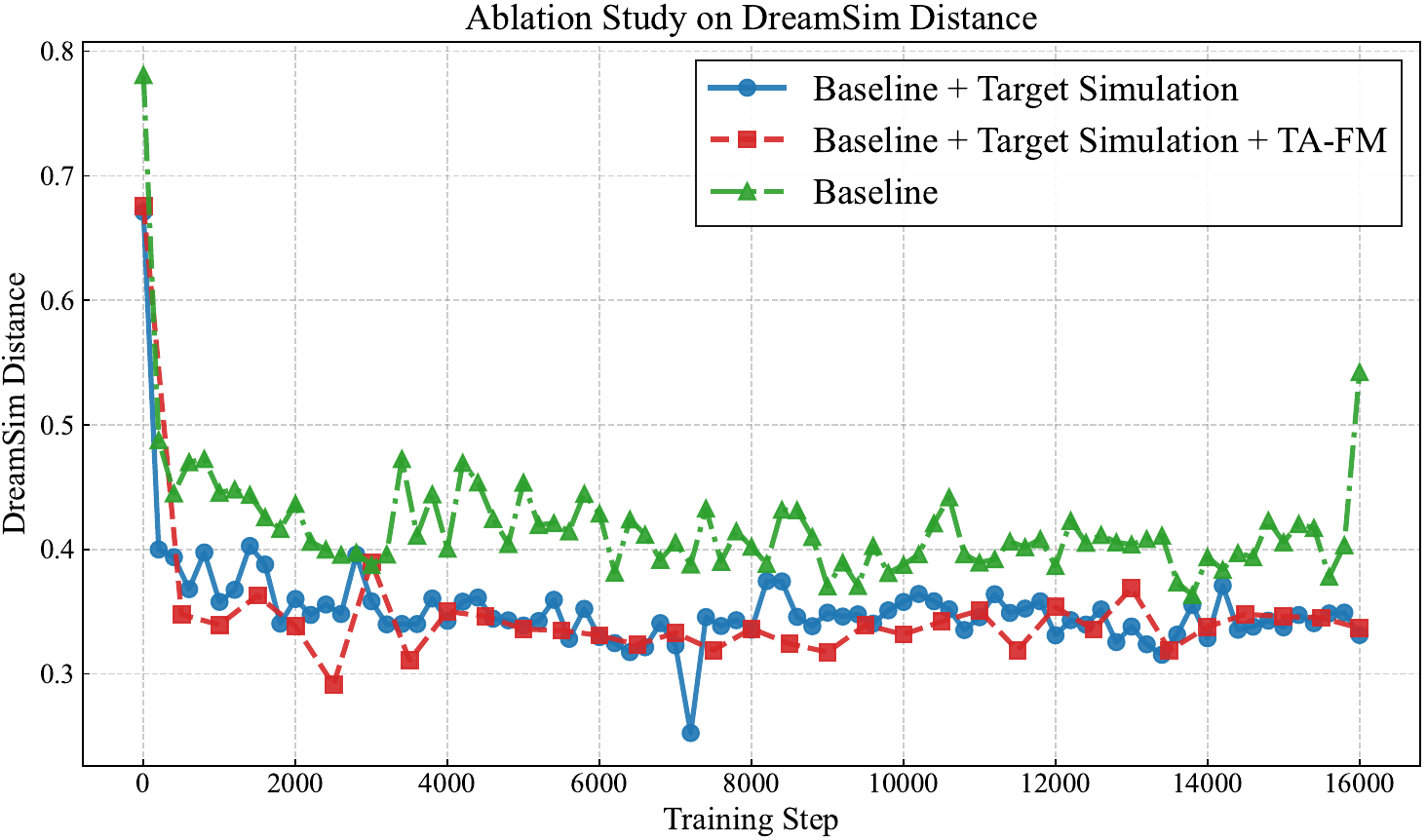}
        \caption{DreamSim Distance}
        \label{fig:ablation_dreamsim}
    \end{subfigure}
    \hfill 
    \begin{subfigure}{0.48\linewidth}
        \centering
        \includegraphics[width=\linewidth]{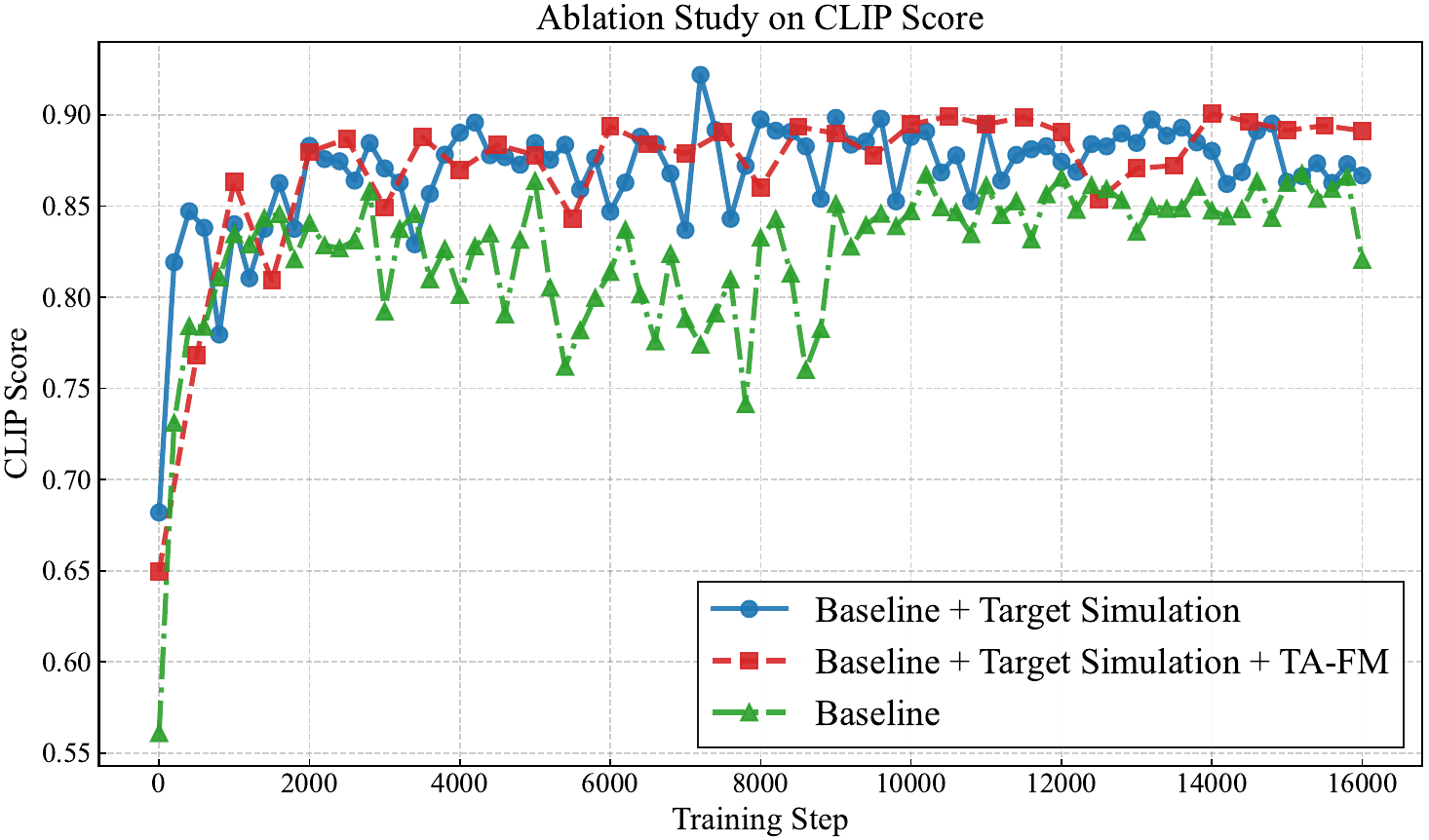}
        \caption{CLIP Score}
        \label{fig:ablation_clip}
    \end{subfigure}
    \caption{\textbf{Quantitative ablation of training dynamics.} The integration of TS and TA-FM significantly accelerates convergence and stabilizes the optimization trajectory compared to the fluctuating baseline.}
    \label{fig:clip_dreamsim_step}
\end{figure}

\noindent
\textbf{Training Dynamics and Stability.}
As shown in Fig.~\ref{fig:clip_dreamsim_step},  the full configuration yields exceptional training stability. The baseline suffers from severe fluctuations under complex multi-task distributions. While TS accelerates early convergence, its trajectory remains volatile. Incorporating TA-FM effectively smooths these oscillations, producing the most stable convergence curve. Synergistically, TS enhances fitting efficiency while TA-FM acts as a crucial stabilizer, ensuring rapid and steady convergence.

\section{Conclusion}
\label{sec:conclusion}
We propose CollectionLoRA, a unified multi-teacher distillation framework that integrates diverse customized visual effects and few-step inference into a single module, eliminating the storage overhead and concept interference (e.g., semantic drift) inherent in traditional multi-LoRA deployments. To address training instability in few-shot multi-concept distillation, our framework features three components: Probabilistic Dual-Stream Routing (PDSR) for structural regularization, Asymmetric Orthogonal Prompting  (AOP) for latent concept isolation, and a Coarse-to-Fine Distillation Objective (C2F-DO) to stabilize optimization and restore high-frequency details. Extensive experiments demonstrate that CollectionLoRA achieves superior concept fidelity, feature isolation, and high-quality generation surpassing single-task teachers, all while maintaining few-step generation capabilities.

\clearpage


%
%
\bibliographystyle{splncs04}
\bibliography{main}

@String(CVPR  = {IEEE Conf. Comput. Vis. Pattern Recog.})

@String(ICLR  = {Int. Conf. Learn. Represent.})

@String(AAAI  = {AAAI})

@String(CVPR  = {CVPR})

@String(ICLR  = {ICLR})

@inproceedings{liu2025tfcustom,
  title={TFCustom: Customized Image Generation with Time-Aware Frequency Feature Guidance},
  author={Liu, Mushui and She, Dong and Pang, Jingxuan and Huang, Qihan and Ying, Jiacheng and He, Wanggui and Hou, Yuanlei and Fu, Siming},
  booktitle={CVPR},
  pages={2714--2723},
  year={2025}
}

@inproceedings{liu2025llm4gen,
  title={Llm4gen: Leveraging semantic representation of llms for text-to-image generation},
  author={Liu, Mushui and Ma, Yuhang and Yang, Zhen and Dan, Jun and Yu, Yunlong and Zhao, Zeng and Hu, Zhipeng and Liu, Bai and Fan, Changjie},
  booktitle={AAAI},
  pages={5523--5531},
  year={2025}
}

@inproceedings{she2025mosaic,
  title={MOSAIC: Multi-Subject Personalized Generation via Correspondence-Aware Alignment and Disentanglement},
  author={She, Dong and Fu, Siming and Liu, Mushui and Jin, Qiaoqiao and Wang, Hualiang and Liu, Mu and Jiang, Jidong},
  booktitle={ICLR},
  year={2026}
}

@article{gal2022image,
  title={An image is worth one word: Personalizing text-to-image generation using textual inversion},
  author={Gal, Rinon and Alaluf, Yuval and Atzmon, Yuval and Patashnik, Or and Bermano, Amit H and Chechik, Gal and Cohen-Or, Daniel},
  journal={arXiv preprint arXiv:2208.01618},
  year={2022}
}

@inproceedings{ruiz2023dreambooth,
  title={Dreambooth: Fine tuning text-to-image diffusion models for subject-driven generation},
  author={Ruiz, Nataniel and Li, Yuanzhen and Jampani, Varun and Pritch, Yael and Rubinstein, Michael and Aberman, Kfir},
  booktitle={Proceedings of the IEEE/CVF conference on computer vision and pattern recognition},
  pages={22500--22510},
  year={2023}
}

@inproceedings{wei2023elite,
  title={Elite: Encoding visual concepts into textual embeddings for customized text-to-image generation},
  author={Wei, Yuxiang and Zhang, Yabo and Ji, Zhilong and Bai, Jinfeng and Zhang, Lei and Zuo, Wangmeng},
  booktitle={Proceedings of the IEEE/CVF international conference on computer vision},
  pages={15943--15953},
  year={2023}
}

@article{ye2023ip,
  title={Ip-adapter: Text compatible image prompt adapter for text-to-image diffusion models},
  author={Ye, Hu and Zhang, Jun and Liu, Sibo and Han, Xiao and Yang, Wei},
  journal={arXiv preprint arXiv:2308.06721},
  year={2023}
}

@article{wang2024instantid,
  title={Instantid: Zero-shot identity-preserving generation in seconds},
  author={Wang, Qixun and Bai, Xu and Wang, Haofan and Qin, Zekui and Chen, Anthony and Li, Huaxia and Tang, Xu and Hu, Yao},
  journal={arXiv preprint arXiv:2401.07519},
  year={2024}
}

@inproceedings{song2024moma,
  title={Moma: Multimodal llm adapter for fast personalized image generation},
  author={Song, Kunpeng and Zhu, Yizhe and Liu, Bingchen and Yan, Qing and Elgammal, Ahmed and Yang, Xiao},
  booktitle={European Conference on Computer Vision},
  pages={117--132},
  year={2024},
  organization={Springer}
}

@inproceedings{kumari2023multi,
  title={Multi-concept customization of text-to-image diffusion},
  author={Kumari, Nupur and Zhang, Bingliang and Zhang, Richard and Shechtman, Eli and Zhu, Jun-Yan},
  booktitle={Proceedings of the IEEE/CVF conference on computer vision and pattern recognition},
  pages={1931--1941},
  year={2023}
}

@inproceedings{mou2025dreamo,
  title={Dreamo: A unified framework for image customization},
  author={Mou, Chong and Wu, Yanze and Wu, Wenxu and Guo, Zinan and Zhang, Pengze and Cheng, Yufeng and Luo, Yiming and Ding, Fei and Zhang, Shiwen and Li, Xinghui and others},
  booktitle={Proceedings of the SIGGRAPH Asia 2025 Conference Papers},
  pages={1--12},
  year={2025}
}

@inproceedings{peebles2023scalable,
  title={Scalable diffusion models with transformers},
  author={Peebles, William and Xie, Saining},
  booktitle={Proceedings of the IEEE/CVF international conference on computer vision},
  pages={4195--4205},
  year={2023}
}

@misc{flux2024,
    author={Black Forest Labs},
    title={FLUX},
    year={2024},
    howpublished={\url{https://github.com/black-forest-labs/flux}},
}

@inproceedings{esser2024scaling,
  title={Scaling rectified flow transformers for high-resolution image synthesis},
  author={Esser, Patrick and Kulal, Sumith and Blattmann, Andreas and Entezari, Rahim and M{\"u}ller, Jonas and Saini, Harry and Levi, Yam and Lorenz, Dominik and Sauer, Axel and Boesel, Frederic and others},
  booktitle={Forty-first international conference on machine learning},
  year={2024}
}

@article{xie2023omnicontrol,
  title={Omnicontrol: Control any joint at any time for human motion generation},
  author={Xie, Yiming and Jampani, Varun and Zhong, Lei and Sun, Deqing and Jiang, Huaizu},
  journal={arXiv preprint arXiv:2310.08580},
  year={2023}
}

@inproceedings{zhang2025easycontrol,
  title={Easycontrol: Adding efficient and flexible control for diffusion transformer},
  author={Zhang, Yuxuan and Yuan, Yirui and Song, Yiren and Wang, Haofan and Liu, Jiaming},
  booktitle={Proceedings of the IEEE/CVF International Conference on Computer Vision},
  pages={19513--19524},
  year={2025}
}

@article{deng2025emerging,
  title={Emerging properties in unified multimodal pretraining},
  author={Deng, Chaorui and Zhu, Deyao and Li, Kunchang and Gou, Chenhui and Li, Feng and Wang, Zeyu and Zhong, Shu and Yu, Weihao and Nie, Xiaonan and Song, Ziang and others},
  journal={arXiv preprint arXiv:2505.14683},
  year={2025}
}

@article{wu2025qwen,
  title={Qwen-image technical report},
  author={Wu, Chenfei and Li, Jiahao and Zhou, Jingren and Lin, Junyang and Gao, Kaiyuan and Yan, Kun and Yin, Sheng-ming and Bai, Shuai and Xu, Xiao and Chen, Yilei and others},
  journal={arXiv preprint arXiv:2508.02324},
  year={2025}
}

@article{labs2025flux,
  title={FLUX. 1 Kontext: Flow Matching for In-Context Image Generation and Editing in Latent Space},
  author={Labs, Black Forest and Batifol, Stephen and Blattmann, Andreas and Boesel, Frederic and Consul, Saksham and Diagne, Cyril and Dockhorn, Tim and English, Jack and English, Zion and Esser, Patrick and others},
  journal={arXiv preprint arXiv:2506.15742},
  year={2025}
}

@misc{flux-2-2025,
    author={Black Forest Labs},
    title={{FLUX.2: Frontier Visual Intelligence}},
    year={2025},
    howpublished={\url{https://bfl.ai/blog/flux-2}},
}

@article{luo2023latent,
  title={Latent consistency models: Synthesizing high-resolution images with few-step inference},
  author={Luo, Simian and Tan, Yiqin and Huang, Longbo and Li, Jian and Zhao, Hang},
  journal={arXiv preprint arXiv:2310.04378},
  year={2023}
}

@misc{lightx2v,
 author = {LightX2V Contributors},
 title = {LightX2V: Light Video Generation Inference Framework},
 year = {2025},
 publisher = {GitHub},
 journal = {GitHub repository},
 howpublished = {\url{https://github.com/ModelTC/lightx2v}},
}

@misc{aitookit,
 author = {aitookit Contributors},
 title = {aitookit},
 year = {2025},
 publisher = {GitHub},
 journal = {GitHub repository},
 howpublished = {\url{https://github.com/ostris/ai-toolkit}},
}

@article{song2023consistency,
  title={Consistency models},
  author={Song, Yang and Dhariwal, Prafulla and Chen, Mark and Sutskever, Ilya},
  year={2023}
}

@article{wang2024phased,
  title={Phased consistency models},
  author={Wang, Fu-Yun and Huang, Zhaoyang and Bergman, Alexander and Shen, Dazhong and Gao, Peng and Lingelbach, Michael and Sun, Keqiang and Bian, Weikang and Song, Guanglu and Liu, Yu and others},
  journal={Advances in neural information processing systems},
  volume={37},
  pages={83951--84009},
  year={2024}
}

@article{zhai2024motion,
  title={Motion consistency model: Accelerating video diffusion with disentangled motion-appearance distillation},
  author={Zhai, Yuanhao and Lin, Kevin and Yang, Zhengyuan and Li, Linjie and Wang, Jianfeng and Lin, Chung-Ching and Doermann, David and Yuan, Junsong and Wang, Lijuan},
  journal={Advances in Neural Information Processing Systems},
  volume={37},
  pages={111000--111021},
  year={2024}
}

@inproceedings{dmd1,
  title={One-step diffusion with distribution matching distillation},
  author={Yin, Tianwei and Gharbi, Micha{\"e}l and Zhang, Richard and Shechtman, Eli and Durand, Fredo and Freeman, William T and Park, Taesung},
  booktitle={Proceedings of the IEEE/CVF conference on computer vision and pattern recognition},
  pages={6613--6623},
  year={2024}
}

@article{dmd2,
  title={Improved distribution matching distillation for fast image synthesis},
  author={Yin, Tianwei and Gharbi, Micha{\"e}l and Park, Taesung and Zhang, Richard and Shechtman, Eli and Durand, Fredo and Freeman, Bill},
  journal={Advances in neural information processing systems},
  volume={37},
  pages={47455--47487},
  year={2024}
}

@article{chen2025flash,
  title={Flash-DMD: Towards High-Fidelity Few-Step Image Generation with Efficient Distillation and Joint Reinforcement Learning},
  author={Chen, Guanjie and Huang, Shirui and Liu, Kai and Zhu, Jianchen and Qu, Xiaoye and Chen, Peng and Cheng, Yu and Sun, Yifu},
  journal={arXiv preprint arXiv:2511.20549},
  year={2025}
}

@article{decoupledmd,
  title={Decoupled DMD: CFG Augmentation as the Spear, Distribution Matching as the Shield},
  author={Liu, Dongyang and Gao, Peng and Liu, David and Du, Ruoyi and Li, Zhen and Wu, Qilong and Jin, Xin and Cao, Sihan and Zhang, Shifeng and Li, Hongsheng and others},
  journal={arXiv preprint arXiv:2511.22677},
  year={2025}
}

@article{jiang2025distribution,
  title={Distribution Matching Distillation Meets Reinforcement Learning},
  author={Jiang, Dengyang and Liu, Dongyang and Wang, Zanyi and Wu, Qilong and Li, Liuzhuozheng and Li, Hengzhuang and Jin, Xin and Liu, David and Li, Zhen and Zhang, Bo and others},
  journal={arXiv preprint arXiv:2511.13649},
  year={2025}
}

@inproceedings{sd1,
  title={High-resolution image synthesis with latent diffusion models},
  author={Rombach, Robin and Blattmann, Andreas and Lorenz, Dominik and Esser, Patrick and Ommer, Bj{\"o}rn},
  booktitle={Proceedings of the IEEE/CVF conference on computer vision and pattern recognition},
  pages={10684--10695},
  year={2022}
}

@article{qwenimage,
  title={Qwen-image technical report},
  author={Wu, Chenfei and Li, Jiahao and Zhou, Jingren and Lin, Junyang and Gao, Kaiyuan and Yan, Kun and Yin, Sheng-ming and Bai, Shuai and Xu, Xiao and Chen, Yilei and others},
  journal={arXiv preprint arXiv:2508.02324},
  year={2025}
}

@misc{zhang2023addingconditionalcontroltexttoimage,
      title={Adding Conditional Control to Text-to-Image Diffusion Models}, 
      author={Lvmin Zhang and Anyi Rao and Maneesh Agrawala},
      year={2023},
      eprint={2302.05543},
      archivePrefix={arXiv},
      primaryClass={cs.CV},
      url={https://arxiv.org/abs/2302.05543}, 
}

@misc{LoRA,
      title={LoRA: Low-Rank Adaptation of Large Language Models}, 
      author={Edward J. Hu and Yelong Shen and Phillip Wallis and Zeyuan Allen-Zhu and Yuanzhi Li and Shean Wang and Lu Wang and Weizhu Chen},
      year={2021},
      eprint={2106.09685},
      archivePrefix={arXiv},
      primaryClass={cs.CL},
      url={https://arxiv.org/abs/2106.09685}, 
}

@misc{OmniConsistency,
      title={OmniConsistency: Learning Style-Agnostic Consistency from Paired Stylization Data}, 
      author={Yiren Song and Cheng Liu and Mike Zheng Shou},
      year={2025},
      eprint={2505.18445},
      archivePrefix={arXiv},
      primaryClass={cs.CV},
      url={https://arxiv.org/abs/2505.18445}, 
}

@article{Photodoodle,
  title={Photodoodle: Learning artistic image editing from few-shot pairwise data},
  author={Huang, Shijie and Song, Yiren and Zhang, Yuxuan and Guo, Hailong and Wang, Xueyin and Shou, Mike Zheng and Liu, Jiaming},
  journal={arXiv preprint arXiv:2502.14397},
  year={2025}
}

@inproceedings{guo2025any2anytryon,
  title={Any2anytryon: Leveraging adaptive position embeddings for versatile virtual clothing tasks},
  author={Guo, Hailong and Zeng, Bohan and Song, Yiren and Zhang, Wentao and Liu, Jiaming and Zhang, Chuang},
  booktitle={Proceedings of the IEEE/CVF International Conference on Computer Vision},
  pages={19085--19096},
  year={2025}
}

@misc{clip,
      title={Learning Transferable Visual Models From Natural Language Supervision}, 
      author={Alec Radford and Jong Wook Kim and Chris Hallacy and Aditya Ramesh and Gabriel Goh and Sandhini Agarwal and Girish Sastry and Amanda Askell and Pamela Mishkin and Jack Clark and Gretchen Krueger and Ilya Sutskever},
      year={2021},
      eprint={2103.00020},
      archivePrefix={arXiv},
      primaryClass={cs.CV},
      url={https://arxiv.org/abs/2103.00020}, 
}

@misc{fu2023dreamsimlearningnewdimensions,
      title={DreamSim: Learning New Dimensions of Human Visual Similarity using Synthetic Data}, 
      author={Stephanie Fu and Netanel Tamir and Shobhita Sundaram and Lucy Chai and Richard Zhang and Tali Dekel and Phillip Isola},
      year={2023},
      eprint={2306.09344},
      archivePrefix={arXiv},
      primaryClass={cs.CV},
      url={https://arxiv.org/abs/2306.09344}, 
}

@misc{wu2026editrewardhumanalignedrewardmodel,
      title={EditReward: A Human-Aligned Reward Model for Instruction-Guided Image Editing}, 
      author={Keming Wu and Sicong Jiang and Max Ku and Ping Nie and Minghao Liu and Wenhu Chen},
      year={2026},
      eprint={2509.26346},
      archivePrefix={arXiv},
      primaryClass={cs.CV},
      url={https://arxiv.org/abs/2509.26346}, 
}

@misc{zhang2022dinodetrimproveddenoising,
      title={DINO: DETR with Improved DeNoising Anchor Boxes for End-to-End Object Detection}, 
      author={Hao Zhang and Feng Li and Shilong Liu and Lei Zhang and Hang Su and Jun Zhu and Lionel M. Ni and Heung-Yeung Shum},
      year={2022},
      eprint={2203.03605},
      archivePrefix={arXiv},
      primaryClass={cs.CV},
      url={https://arxiv.org/abs/2203.03605}, 
}

@misc{wu2025dcoardeepconceptinjection,
      title={DCoAR: Deep Concept Injection into Unified Autoregressive Models for Personalized Text-to-Image Generation}, 
      author={Fangtai Wu and Mushui Liu and Weijie He and Zhao Wang and Yunlong Yu},
      year={2025},
      eprint={2508.07341},
      archivePrefix={arXiv},
      primaryClass={cs.CV},
      url={https://arxiv.org/abs/2508.07341}, 
}

@misc{lipman2023flowmatchinggenerativemodeling,
      title={Flow Matching for Generative Modeling}, 
      author={Yaron Lipman and Ricky T. Q. Chen and Heli Ben-Hamu and Maximilian Nickel and Matt Le},
      year={2023},
      eprint={2210.02747},
      archivePrefix={arXiv},
      primaryClass={cs.LG},
      url={https://arxiv.org/abs/2210.02747}, 
}

@misc{huang2024incontextloradiffusiontransformers,
      title={In-Context LoRA for Diffusion Transformers}, 
      author={Lianghua Huang and Wei Wang and Zhi-Fan Wu and Yupeng Shi and Huanzhang Dou and Chen Liang and Yutong Feng and Yu Liu and Jingren Zhou},
      year={2024},
      eprint={2410.23775},
      archivePrefix={arXiv},
      primaryClass={cs.CV},
      url={https://arxiv.org/abs/2410.23775}, 
}

@misc{yang2025qwen3technicalreport,
      title={Qwen3 Technical Report}, 
      author={An Yang and Anfeng Li and Baosong Yang and Beichen Zhang and Binyuan Hui and Bo Zheng and Bowen Yu and Chang Gao and Chengen Huang and Chenxu Lv and Chujie Zheng and Dayiheng Liu and Fan Zhou and Fei Huang and Feng Hu and Hao Ge and Haoran Wei and Huan Lin and Jialong Tang and Jian Yang and Jianhong Tu and Jianwei Zhang and Jianxin Yang and Jiaxi Yang and Jing Zhou and Jingren Zhou and Junyang Lin and Kai Dang and Keqin Bao and Kexin Yang and Le Yu and Lianghao Deng and Mei Li and Mingfeng Xue and Mingze Li and Pei Zhang and Peng Wang and Qin Zhu and Rui Men and Ruize Gao and Shixuan Liu and Shuang Luo and Tianhao Li and Tianyi Tang and Wenbiao Yin and Xingzhang Ren and Xinyu Wang and Xinyu Zhang and Xuancheng Ren and Yang Fan and Yang Su and Yichang Zhang and Yinger Zhang and Yu Wan and Yuqiong Liu and Zekun Wang and Zeyu Cui and Zhenru Zhang and Zhipeng Zhou and Zihan Qiu},
      year={2025},
      eprint={2505.09388},
      archivePrefix={arXiv},
      primaryClass={cs.CL},
      url={https://arxiv.org/abs/2505.09388}, 
}

@misc{coreteam2026mimov2flashtechnicalreport,
      title={MiMo-V2-Flash Technical Report}, 
      author={Core Team and Bangjun Xiao and Bingquan Xia and Bo Yang and Bofei Gao and Bowen Shen and Chen Zhang and Chenhong He and Chiheng Lou and Fuli Luo and Gang Wang and Gang Xie and Hailin Zhang and Hanglong Lv and Hanyu Li and Heyu Chen and Hongshen Xu and Houbin Zhang and Huaqiu Liu and Jiangshan Duo and Jianyu Wei and Jiebao Xiao and Jinhao Dong and Jun Shi and Junhao Hu and Kainan Bao and Kang Zhou and Lei Li and Liang Zhao and Linghao Zhang and Peidian Li and Qianli Chen and Shaohui Liu and Shihua Yu and Shijie Cao and Shimao Chen and Shouqiu Yu and Shuo Liu and Tianling Zhou and Weijiang Su and Weikun Wang and Wenhan Ma and Xiangwei Deng and Bohan Mao and Bowen Ye and Can Cai and Chenghua Wang and Chengxuan Zhu and Chong Ma and Chun Chen and Chunan Li and Dawei Zhu and Deshan Xiao and Dong Zhang and Duo Zhang and Fangyue Liu and Feiyu Yang and Fengyuan Shi and Guoan Wang and Hao Tian and Hao Wu and Heng Qu and Hongfei Yi and Hongxu An and Hongyi Guan and Xing Zhang and Yifan Song and Yihan Yan and Yihao Zhao and Yingchun Lai and Yizhao Gao and Yu Cheng and Yuanyuan Tian and Yudong Wang and Zhen Tang and Zhengju Tang and Zhengtao Wen and Zhichao Song and Zhixian Zheng and Zihan Jiang and Jian Wen and Jiarui Sun and Jiawei Li and Jinlong Xue and Jun Xia and Kai Fang and Menghang Zhu and Nuo Chen and Qian Tu and Qihao Zhang and Qiying Wang and Rang Li and Rui Ma and Shaolei Zhang and Shengfan Wang and Shicheng Li and Shuhao Gu and Shuhuai Ren and Sirui Deng and Tao Guo and Tianyang Lu and Weiji Zhuang and Weikang Zhang and Weimin Xiong and Wenshan Huang and Wenyu Yang and Xin Zhang and Xing Yong and Xu Wang and Xueyang Xie and Yilin Jiang and Yixin Yang and Yongzhe He and Yu Tu and Yuanliang Dong and Yuchen Liu and Yue Ma and Yue Yu and Yuxing Xiang and Zhaojun Huang and Zhenru Lin and Zhipeng Xu and Zhiyang Chen and Zhonghua Deng and Zihan Zhang and Zihao Yue},
      year={2026},
      eprint={2601.02780},
      archivePrefix={arXiv},
      primaryClass={cs.CL},
      url={https://arxiv.org/abs/2601.02780}, 
}

@misc{deepseek_v4_pro,
  author       = {{DeepSeek-AI}},
  title        = {DeepSeek-V4-Pro Model Card},
  year         = {2026}, 
  howpublished = {\url{https://huggingface.co/deepseek-ai/DeepSeek-V4-Pro}},
  publisher    = {Hugging Face},
  note         = {Accessed: 2026-05-04} 
}

@misc{llmopd,
      title={On-Policy Distillation of Language Models: Learning from Self-Generated Mistakes}, 
      author={Rishabh Agarwal and Nino Vieillard and Yongchao Zhou and Piotr Stanczyk and Sabela Ramos and Matthieu Geist and Olivier Bachem},
      year={2024},
      eprint={2306.13649},
      archivePrefix={arXiv},
      primaryClass={cs.LG},
      url={https://arxiv.org/abs/2306.13649}, 
}

@misc{minillm,
      title={MiniLLM: On-Policy Distillation of Large Language Models}, 
      author={Yuxian Gu and Li Dong and Furu Wei and Minlie Huang},
      year={2026},
      eprint={2306.08543},
      archivePrefix={arXiv},
      primaryClass={cs.CL},
      url={https://arxiv.org/abs/2306.08543}, 
}

@misc{rethinkingopd,
      title={Rethinking On-Policy Distillation of Large Language Models: Phenomenology, Mechanism, and Recipe}, 
      author={Yaxuan Li and Yuxin Zuo and Bingxiang He and Jinqian Zhang and Chaojun Xiao and Cheng Qian and Tianyu Yu and Huan-ang Gao and Wenkai Yang and Zhiyuan Liu and Ning Ding},
      year={2026},
      eprint={2604.13016},
      archivePrefix={arXiv},
      primaryClass={cs.LG},
      url={https://arxiv.org/abs/2604.13016}, 
}

@misc{rlopd,
      title={Learning beyond Teacher: Generalized On-Policy Distillation with Reward Extrapolation}, 
      author={Wenkai Yang and Weijie Liu and Ruobing Xie and Kai Yang and Saiyong Yang and Yankai Lin},
      year={2026},
      eprint={2602.12125},
      archivePrefix={arXiv},
      primaryClass={cs.LG},
      url={https://arxiv.org/abs/2602.12125}, 
}

@misc{fang2026flowopdonpolicydistillationflow,
      title={Flow-OPD: On-Policy Distillation for Flow Matching Models}, 
      author={Zhen Fang and Wenxuan Huang and Yu Zeng and Yiming Zhao and Shuang Chen and Kaituo Feng and Yunlong Lin and Lin Chen and Zehui Chen and Shaosheng Cao and Feng Zhao},
      year={2026},
      eprint={2605.08063},
      archivePrefix={arXiv},
      primaryClass={cs.CV},
      url={https://arxiv.org/abs/2605.08063}, 
}

@misc{jiang2026dopsdonpolicyselfdistillationcontinuously,
      title={D-OPSD: On-Policy Self-Distillation for Continuously Tuning Step-Distilled Diffusion Models}, 
      author={Dengyang Jiang and Xin Jin and Dongyang Liu and Zanyi Wang and Mingzhe Zheng and Ruoyi Du and Xiangpeng Yang and Qilong Wu and Zhen Li and Peng Gao and Harry Yang and Steven Hoi},
      year={2026},
      eprint={2605.05204},
      archivePrefix={arXiv},
      primaryClass={cs.CV},
      url={https://arxiv.org/abs/2605.05204}, 
}

@misc{chern2025livetalkrealtimemultimodalinteractive,
      title={LiveTalk: Real-Time Multimodal Interactive Video Diffusion via Improved On-Policy Distillation}, 
      author={Ethan Chern and Zhulin Hu and Bohao Tang and Jiadi Su and Steffi Chern and Zhijie Deng and Pengfei Liu},
      year={2025},
      eprint={2512.23576},
      archivePrefix={arXiv},
      primaryClass={cs.CV},
      url={https://arxiv.org/abs/2512.23576}, 
}

@misc{gu2026anyflowanystepvideodiffusion,
      title={AnyFlow: Any-Step Video Diffusion Model with On-Policy Flow Map Distillation}, 
      author={Yuchao Gu and Guian Fang and Yuxin Jiang and Weijia Mao and Song Han and Han Cai and Mike Zheng Shou},
      year={2026},
      eprint={2605.13724},
      archivePrefix={arXiv},
      primaryClass={cs.CV},
      url={https://arxiv.org/abs/2605.13724}, 
}

\clearpage
\section*{Overview of Supplementary Material}

This supplementary document provides comprehensive technical details, in-depth theoretical analyses, and extensive visual results to fully support the claims made in the main manuscript. The contents are systematically organized as follows:

\begin{itemize}
    \item Appendix 7: Dataset and Evaluation Protocols.
    \item Appendix 8: Extended Empirical Analyses.
    \item Appendix 9: Implementation Details.
    \item Appendix 10: User Study and Additional Qualitative Results.
\end{itemize}

\section{Dataset and Evaluation Protocols}
\subsection{Dataset Details}
We utilized internally constructed special-effect image pairs for LoRA training. In total, we collected 180 distinct effects, with approximately 20 training pairs for each. 
\subsection{Prompt Design for Evaluation}

\begin{figure}[htbp]
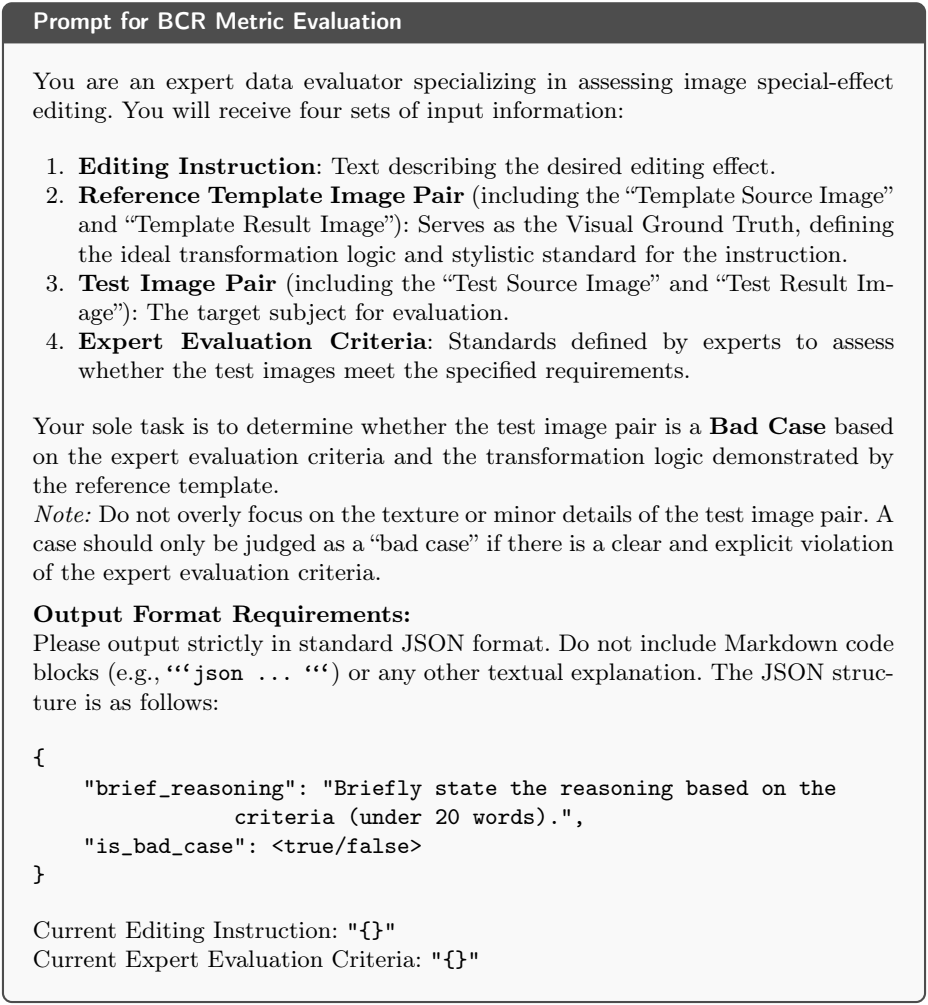

\centering
\begin{tcolorbox}[
    colback=gray!5!white,      
    colframe=gray!60!black,    
    title=\textbf{Prompt for BCR Metric Evaluation}, 
    fonttitle=\sffamily\bfseries,
    boxrule=0.8pt,             
    arc=3pt,                   
    left=8pt, right=8pt, top=8pt, bottom=8pt 
]
\small 

You are an expert data evaluator specializing in assessing image special-effect editing. You will receive four sets of input information:

\begin{enumerate}
    \item \textbf{Editing Instruction}: Text describing the desired editing effect.
    \item \textbf{Reference Template Image Pair} (including the ``Template Source Image'' and ``Template Result Image''): Serves as the Visual Ground Truth, defining the ideal transformation logic and stylistic standard for the instruction.
    \item \textbf{Test Image Pair} (including the ``Test Source Image'' and ``Test Result Image''): The target subject for evaluation.
    \item \textbf{Expert Evaluation Criteria}: Standards defined by experts to assess whether the test images meet the specified requirements.
\end{enumerate}

Your sole task is to determine whether the test image pair is a \textbf{Bad Case} based on the expert evaluation criteria and the transformation logic demonstrated by the reference template. 

\textit{Note:} Do not overly focus on the texture or minor details of the test image pair. A case should only be judged as a ``bad case'' if there is a clear and explicit violation of the expert evaluation criteria.

\vspace{0.5em}
\noindent\textbf{Output Format Requirements:}\\
Please output strictly in standard JSON format. Do not include Markdown code blocks (e.g., \texttt{```json ... ```}) or any other textual explanation. The JSON structure is as follows:

\begin{verbatim}
{
    "brief_reasoning": "Briefly state the reasoning based on the 
                criteria (under 20 words).",
    "is_bad_case": <true/false>
}
\end{verbatim}

\noindent Current Editing Instruction: \texttt{"\{\}"} \\
Current Expert Evaluation Criteria: \texttt{"\{\}"}

\end{tcolorbox}
\caption{The prompt template used for querying the Multimodal Large Language Model (MLLM) to evaluate the BCR metric.}
\label{fig:bcr_prompt}
\end{figure}

We employ a Multimodal Large Language Model (MLLM) to evaluate our VSA and BCR metrics. Specifically, we utilize the \texttt{Qwen-VL-Max-Latest} API for the evaluation of all test samples. To assess the BCR metric, we apply the prompt detailed in Figure~\ref{fig:bcr_prompt}. For the VSA metric, we initially utilize the prompt presented in Figure~\ref{fig:bcr_prompt} to determine whether the generated result constitutes a bad case, and subsequently employ the prompt illustrated in Figure~\ref{fig:vsa_quantitative_prompt} to evaluate the consistency score.

\begin{figure}[htbp]
\centering
\begin{tcolorbox}[
    colback=gray!5!white,
    colframe=gray!60!black,
    title=\textbf{Prompt for VSA (Consistency) Quantitative Evaluation},
    fonttitle=\sffamily\bfseries,
    boxrule=0.8pt,
    arc=3pt,
    left=8pt, right=8pt, top=8pt, bottom=8pt
]
\small

You are an expert data evaluator specializing in assessing image special-effect editing. You will receive two sets of input information:

\begin{enumerate}
    \item \textbf{Editing Instruction}: Text describing the desired editing effect.
    \item \textbf{Test Image Pair} (including the ``Test Source Image'' and ``Test Result Image''): The target subject for evaluation.
\end{enumerate}

Your task is to evaluate whether the core semantic identity of the subject in the source image is preserved after the special effect is applied. The specific evaluation criteria are as follows:

\vspace{0.5em}
\noindent\textit{Note: For heavy stylization or deformation tasks, please base your judgment on ``semantic recognizability'' rather than pixel-level alignment.}

\begin{itemize}
    \item \textbf{Score 1}: The subject is lost or unrecognizable.
    \item \textbf{Score 2}: The subject is severely distorted, with key features lost.
    \item \textbf{Score 3}: The subject is generally recognizable, but there is noticeable loss of non-essential features.
    \item \textbf{Score 4}: Identity is well-preserved; changes are strictly limited to those required by the special effect.
    \item \textbf{Score 5}: The subject's demeanor and core features are perfectly preserved.
\end{itemize}

\vspace{0.5em}
\noindent\textbf{Output Format Requirements:}\\
Please output strictly in standard JSON format. Do not include Markdown code blocks (e.g., \texttt{```json ... ```}) or any other textual explanation. The JSON structure is as follows:

\begin{verbatim}
{
    "brief_reasoning": "Briefly state the reasoning based on the 
        criteria (under 20 words).",
    "scores": <1-5>
}
\end{verbatim}

\noindent Current Editing Instruction: \texttt{"\{\}"}

\end{tcolorbox}
\caption{The prompt template used for querying the MLLM to quantitatively evaluate the consistency score (1-5) of the generated image.}
\label{fig:vsa_quantitative_prompt}
\end{figure}

\begin{figure*}[htb]
    \centering
    \includegraphics[width=\linewidth]{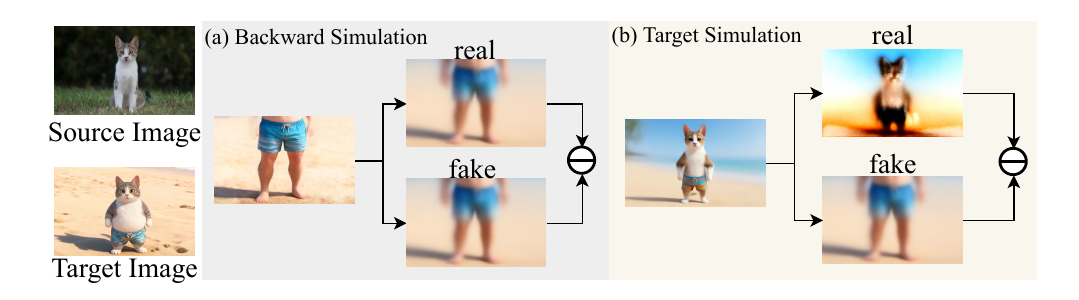}
    \caption{Comparison between (a) Backward Simulation and (b) our proposed Target Simulation. In the heterogeneous distillation setting, Backward Simulation can yield nearly identical 'real' and 'fake' predictions due to severe domain deviation, leading to the vanishing gradient problem. Conversely, Target Simulation provides distinct representations, ensuring informative gradients for the student model.}
    \label{fig:grad_van}
    \vspace{-0.5cm}
\end{figure*}

\section{Extended Empirical Analyses}
\subsection{Analysis of the Vanishing Gradient Problem}
The vanishing gradient problem primarily emerges when the original DMD is directly applied in a heterogeneous distillation setting. In such a setup, the student and teacher models exhibit misaligned distributions, as the student is tasked with acquiring novel capabilities (e.g., special-effect generation) during the training process. In certain extreme cases, the samples generated by the student model via Backward Simulation deviate too drastically from the teacher model's domain. This severe discrepancy causes the teacher model's predictions to be nearly identical to the fake outputs, driving the DMD update gradient close to zero. We refer to this phenomenon as the \textit{vanishing gradient problem in heterogeneous distillation}. The Target Simulation approach proposed in this paper effectively resolves this issue, as illustrated in Figure~\ref{fig:grad_van}.

\subsection{Ablation Study on Timestep-Constrained Target Simulation}

\begin{figure*}[htb]
    \centering
    \includegraphics[width=\linewidth]{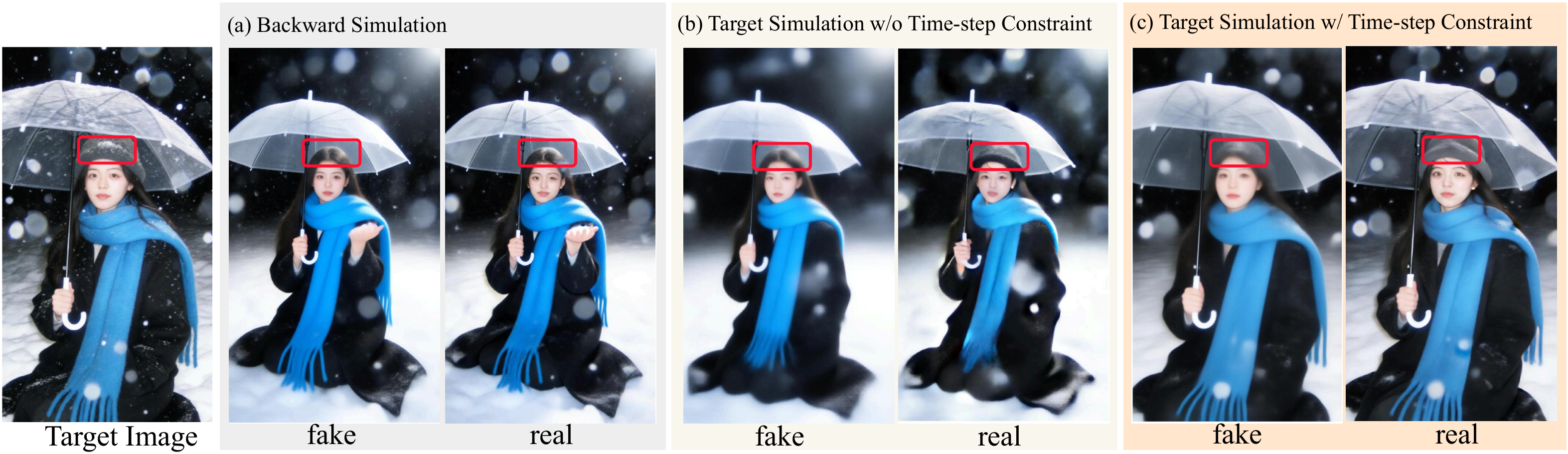}
    \caption{\textbf{Comparison of simulation strategies.} (a) Backward simulation leads to vanishing gradients. (b) Target simulation enables differentiation. (c) Time-step constraints further amplify the discrepancy, providing robust gradient signals for effective training.}
    \label{fig:ablation_timestep}
    \vspace{-0.5cm}
\end{figure*}

Upon the introduction of Target Simulation, we further incorporate a time-step constraint. This refinement aims to amplify the discrepancy between the ``real'' and ``fake'' predictions, thereby facilitating more effective gradient updates for the student model. As illustrated in Figure~\ref{fig:ablation_timestep}, while Target Simulation without the time-step constraint avoids the absolute vanishing gradient problem, it significantly reduces the magnitude of gradient updates compared to the setting where the constraint is imposed.


\section{Implementation Details}
We leverage the \texttt{Qwen-VL-Max-Latest} API to generate student prompts, which not only automates the prompt engineering process but also allows us to fully utilize existing teacher models while mitigating potential prompt conflicts in all-in-one special-effect scenarios. Specifically, we provide the MLLM with two randomly sampled training pairs and the corresponding teacher prompt to generate the refined student prompt. The prompt template is illustrated in Figure~\ref{fig:aac_prompt}, and several illustrative examples are provided in Table~\ref{tab:prompt_comparison}.
\begin{figure}[htbp]
\centering
\begin{tcolorbox}[
    colback=gray!5!white,
    colframe=gray!60!black,
    title=\textbf{Prompt for AAC},
    fonttitle=\sffamily\bfseries,
    boxrule=0.8pt,
    arc=3pt,
    left=8pt, right=8pt, top=8pt, bottom=8pt
]
\small

You are provided with four images, consisting of two pairs of image editing data. In each pair, the first image is the original, and the second is the result generated by an image generation model. This model was trained on such pairs using a generic prompt. Your task is to refine and enrich this prompt to more accurately describe the transformation applied by this effect while maintaining its generality.

\vspace{0.5em}
\noindent\textbf{Constraints and Requirements:}
\begin{enumerate}
    \item \textbf{Subject Generalization}: Do not use specific descriptions for the subjects. Instead, use generic categories (e.g., human, pet, animal) based on the provided images.
    \item \textbf{Perspective Consistency}: Observe the viewpoint/perspective of the editing results. Describe the perspective shift only if it is consistently applied across all pairs; otherwise, omit it.
    \item \textbf{Reference Baseline}: Refer to the provided rough prompt: \texttt{"\{\}"}.
    \item \textbf{Output Specification}: Directly return the refined prompt without any supplementary or conversational text.
\end{enumerate}

\end{tcolorbox}
\caption{The prompt template used for refining and enriching the generic editing prompt based on visual samples and a baseline description.}
\label{fig:aac_prompt}
\end{figure}

\begin{table}[htbp]
\centering
\caption{Examples of Teacher Prompts and the corresponding Refined Student Prompts.}
\label{tab:prompt_comparison}
\footnotesize 
\renewcommand{\arraystretch}{1.3} 
\begin{tabularx}{\textwidth}{@{} l X X @{}}
\toprule
\textbf{LoRA Name} & \textbf{Teacher Prompt} & \textbf{Student Prompt (with Trigword)} \\
\midrule

1117\_3dshatanpangchongwu & 
Change the style to 3D realistic, change the background to a beach, change the pose to standing, and add swim trunks. & 
\textless 1117\_\allowbreak 3dshatanpangchongwu\_\allowbreak Xy5995\textgreater Transform the pet into a 3D realistic style, place it against a beach background, adjust its pose to standing, dress it in brightly colored swim trunks, maintain the fur texture, and keep the overall style cute and expressive. \\

\addlinespace 

1117\_chongwushouzhi & 
Based on the pet in the original image, retain its fur color, patterns, demeanor, expression, and body proportions, shrink it to finger size, and have it lying on a human finger. & 
\textless 1117\_\allowbreak chongwushouzhi\_\allowbreak RN8pcKDi\textgreater Based on the pet in the original image, retain its fur color, patterns, demeanor, expression, and body proportions, reduce it to finger size, place it lying on a human finger, maintain the same perspective, softly blur the background, and enhance realism and fine texture. \\

\bottomrule
\end{tabularx}
\end{table}


\begin{figure}[htbp]
    \centering
    \includegraphics[width=\linewidth]{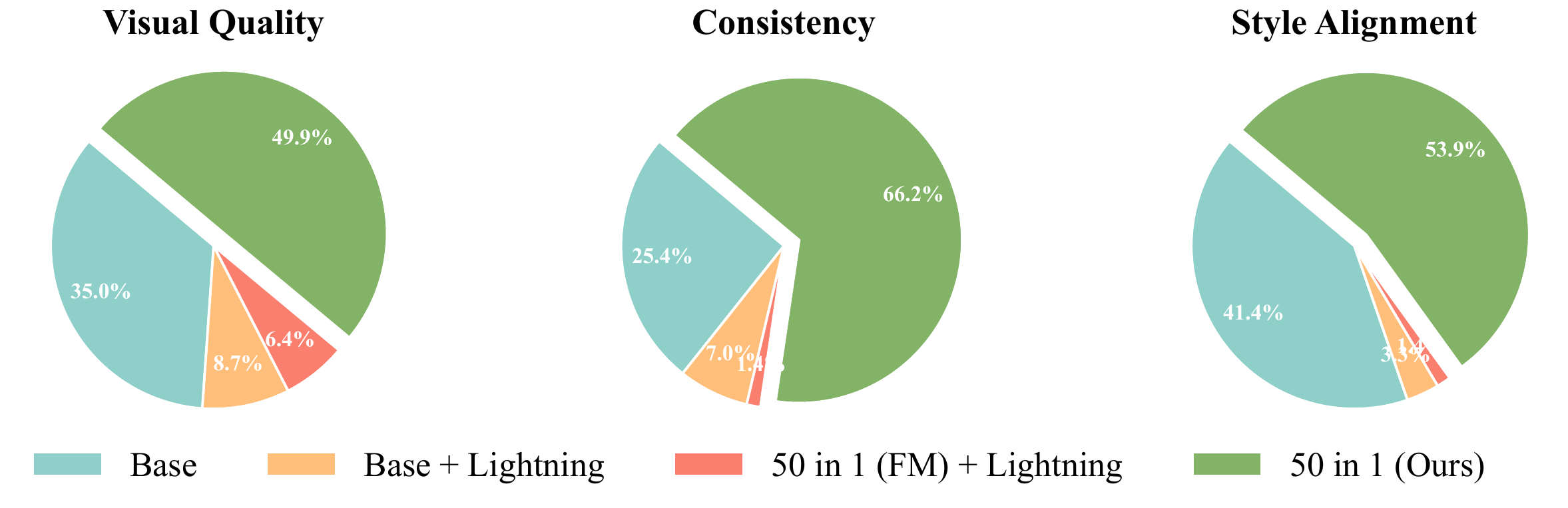}
    \caption{\textbf{Detailed user study results.} Evaluators were asked to choose the best result among four candidates across three dimensions: Visual Quality, Consistency, and Style Alignment. Our proposed method, \textit{50 in 1 (Ours)}, consistently achieves the highest preference in all categories. Most notably, it secures 66.2\% of the votes for Consistency and 53.9\% for Style Alignment, significantly outperforming the \textit{Base} model and the \textit{Lightning} accelerated variants.}
    \label{fig:user_study_supp}
\end{figure}

\section{User Study and Additional Qualitative Results}
To subjectively evaluate our method, we conducted a user study involving 10 professional evaluators. We randomly sampled 50 test sets, where each set comprises the original reference image and four generated candidates: \textit{Base}, \textit{Base+Lightning}, \textit{50 in 1 (FM) + Lightning}, and \textit{Ours}. Evaluators were kept blind to the underlying methods. They were instructed to choose the best image among the four candidates based on three criteria: image quality, consistency, and style alignment. 
As shown in Figure~\ref{fig:user_study_supp}, we present the detailed preference distribution from our user study. Overall, our proposed method, \textit{50 in 1 (Ours)}, consistently outperforms all baseline methods across the three evaluated dimensions according to human perception.
Specifically, for \textbf{Visual Quality}, our method received the highest preference at 49.9\%, demonstrating that evaluators found our generated results to have superior visual appeal. The \textit{Base} model followed as the second most preferred with 35.0\% of the votes. 
In terms of \textbf{Consistency}, the advantage of our method is remarkably pronounced. It was selected as the best by a large margin of 66.2\%, indicating that our approach is highly effective in maintaining consistency compared to the alternatives. 
Regarding \textbf{Style Alignment}, our method again secured the majority of user preference (53.9\%), followed by the \textit{Base} model (41.4\%). Noticeably, the \textit{Lightning} variants (\textit{Base + Lightning} and \textit{50 in 1 (FM) + Lightning}) received minimal votes across all metrics. This suggests that while these variants might offer specific technical advantages, such as inference speed, they generally compromise subjective visual quality and stylistic fidelity. Ultimately, these results strongly validate that our method achieves the best overall balance of visual quality, consistency, and style alignment.

\begin{figure*}[htb]
    \centering
    \includegraphics[width=\linewidth]{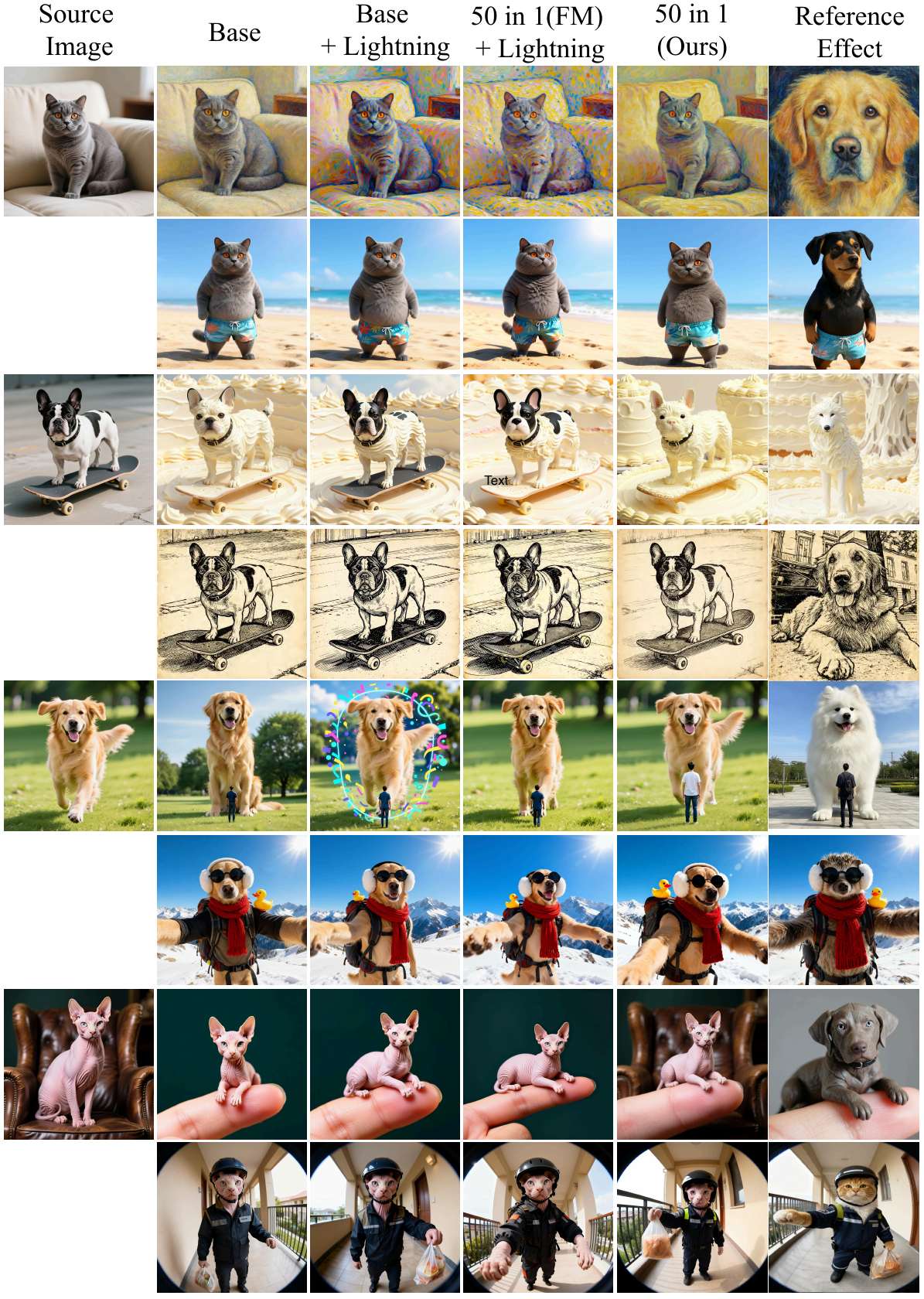}
    \caption{Qualitative Evaluation.}
    \label{fig:visual1}
\end{figure*}

\begin{figure*}[htb]
    \centering
    \includegraphics[width=\linewidth]{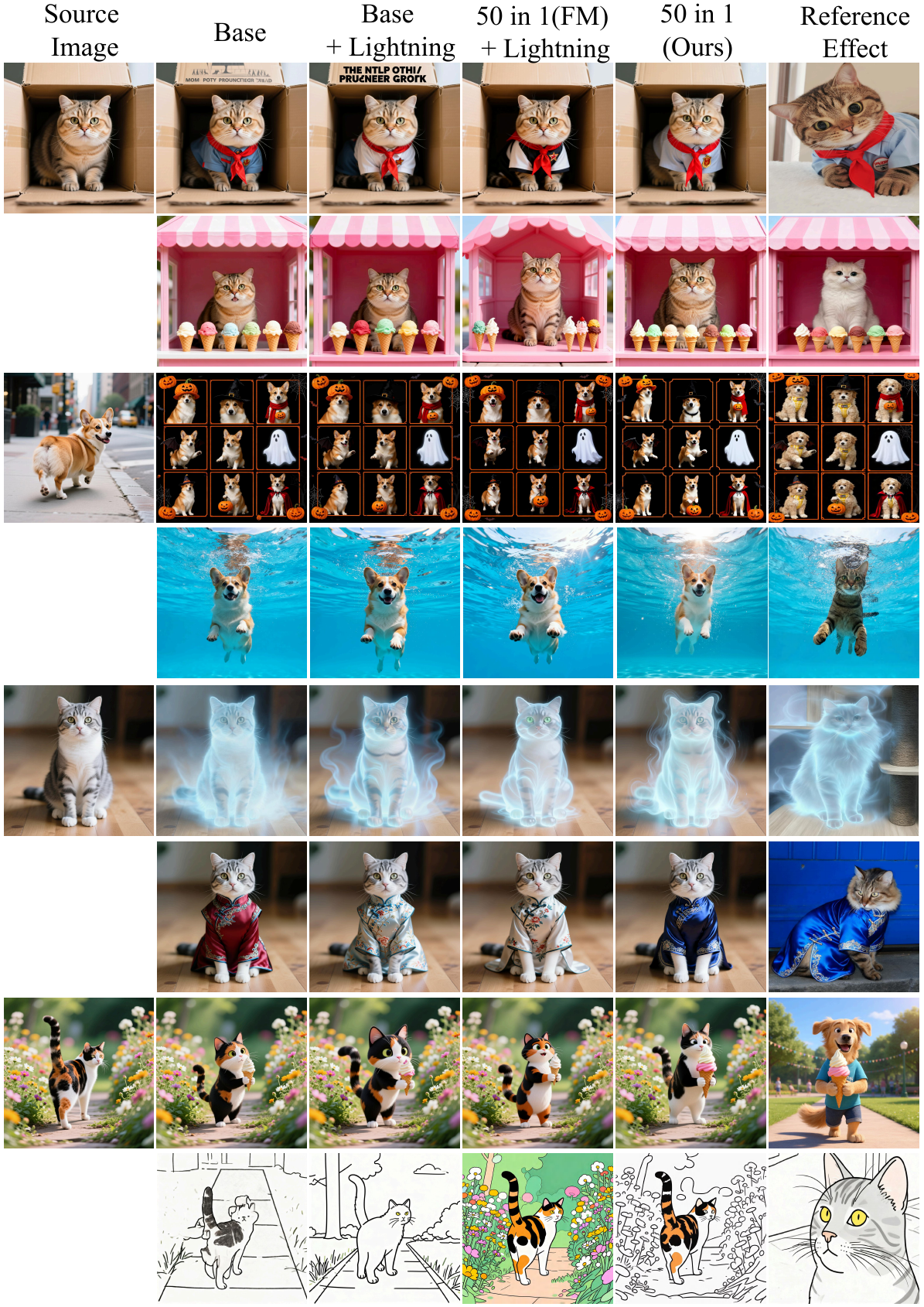}
    \caption{Qualitative Evaluation.}
    \label{fig:visual2}
    \vspace{-0.5cm}
\end{figure*}

\begin{figure*}[htb]
    \centering
    \includegraphics[width=\linewidth]{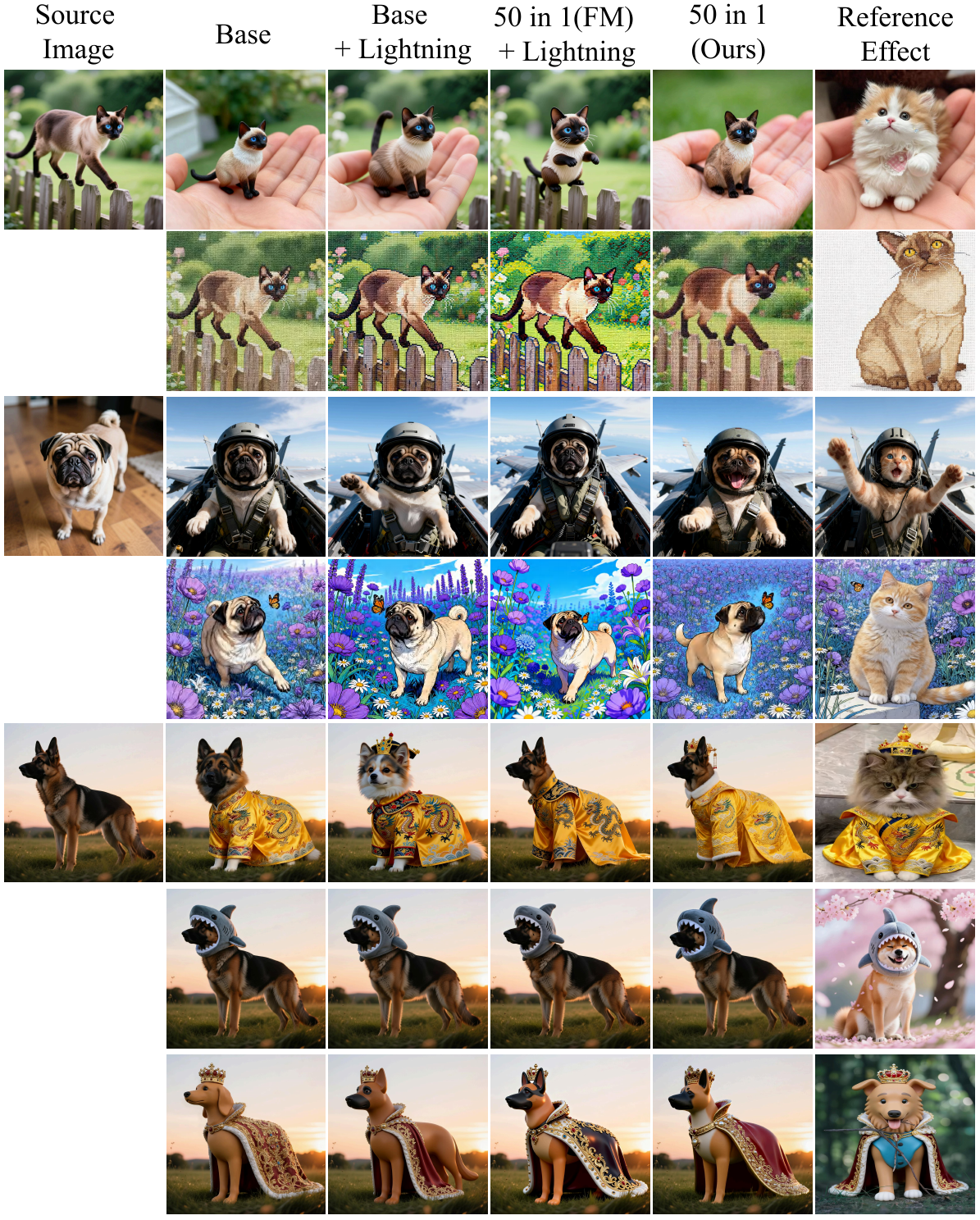}
    \caption{Qualitative Evaluation.}
    \label{fig:visual3}
    \vspace{-0.5cm}
\end{figure*}

\begin{figure*}[htb]
    \centering
    \includegraphics[width=\linewidth]{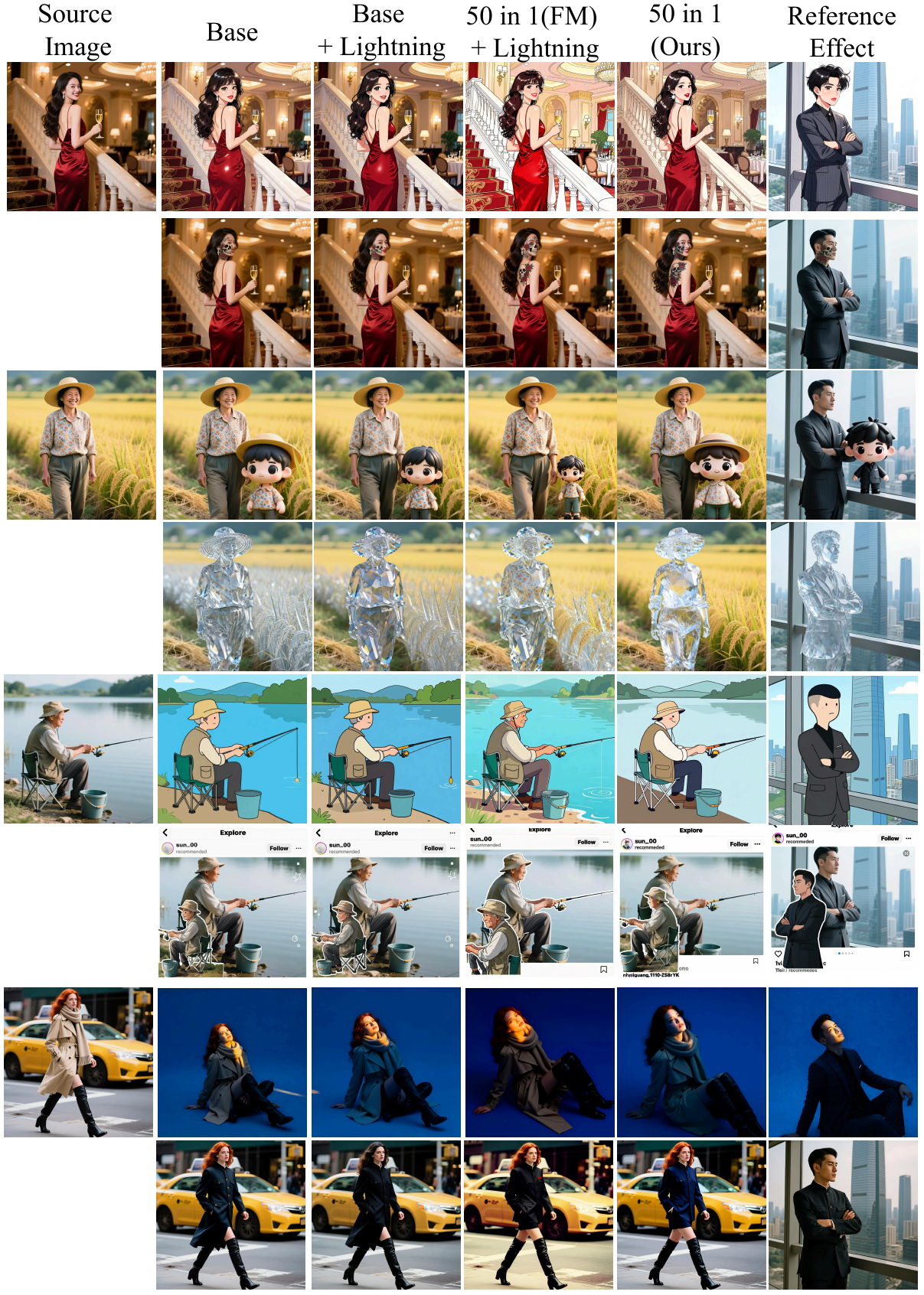}
    \caption{Qualitative Evaluation.}
    \label{fig:visual4}
    \vspace{-0.5cm}
\end{figure*}

\begin{figure*}[htb]
    \centering
    \includegraphics[width=\linewidth]{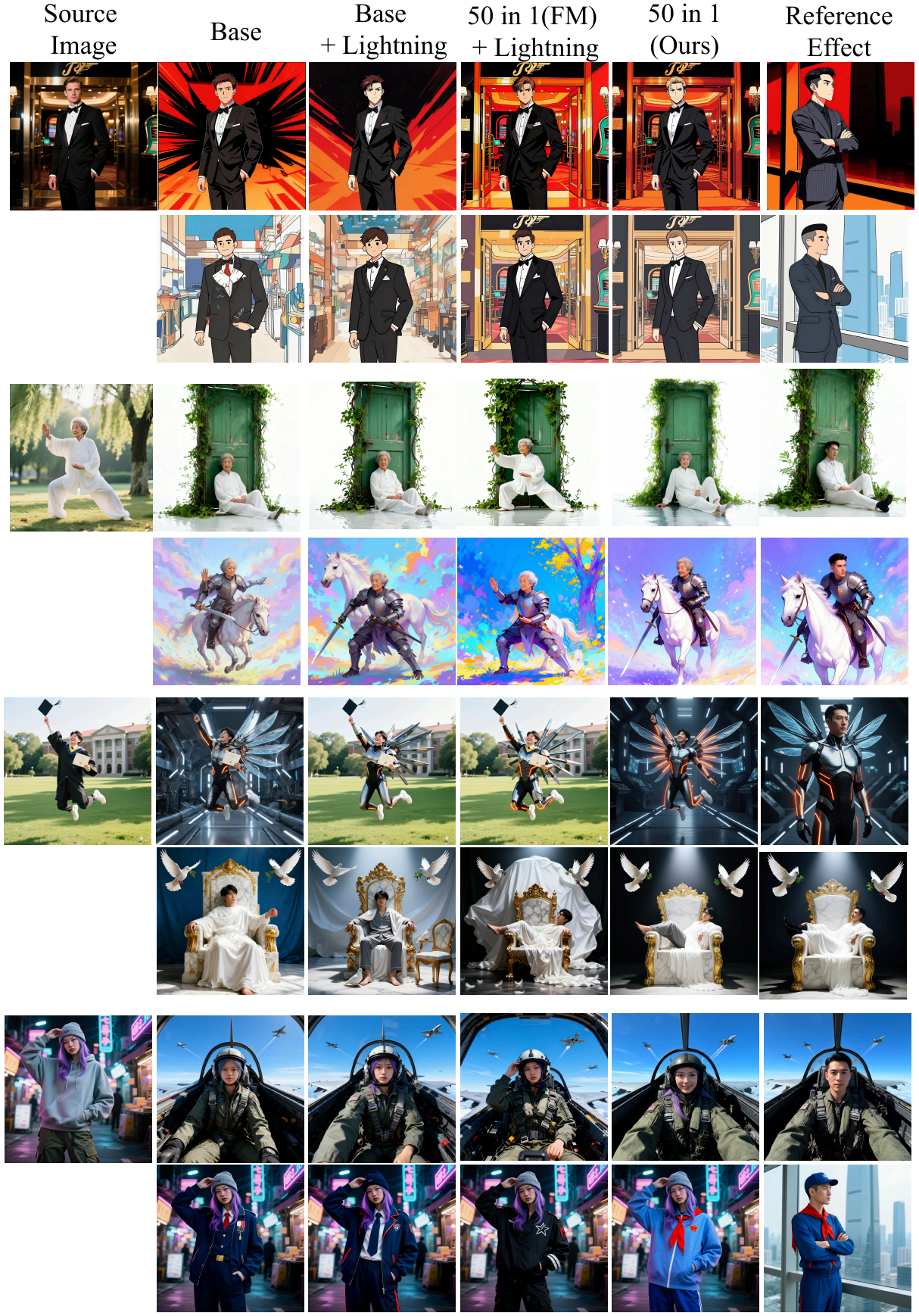}
    \caption{Qualitative Evaluation.}
    \label{fig:visual5}
    \vspace{-0.5cm}
\end{figure*}

\begin{figure*}[htb]
    \centering
    \includegraphics[width=\linewidth]{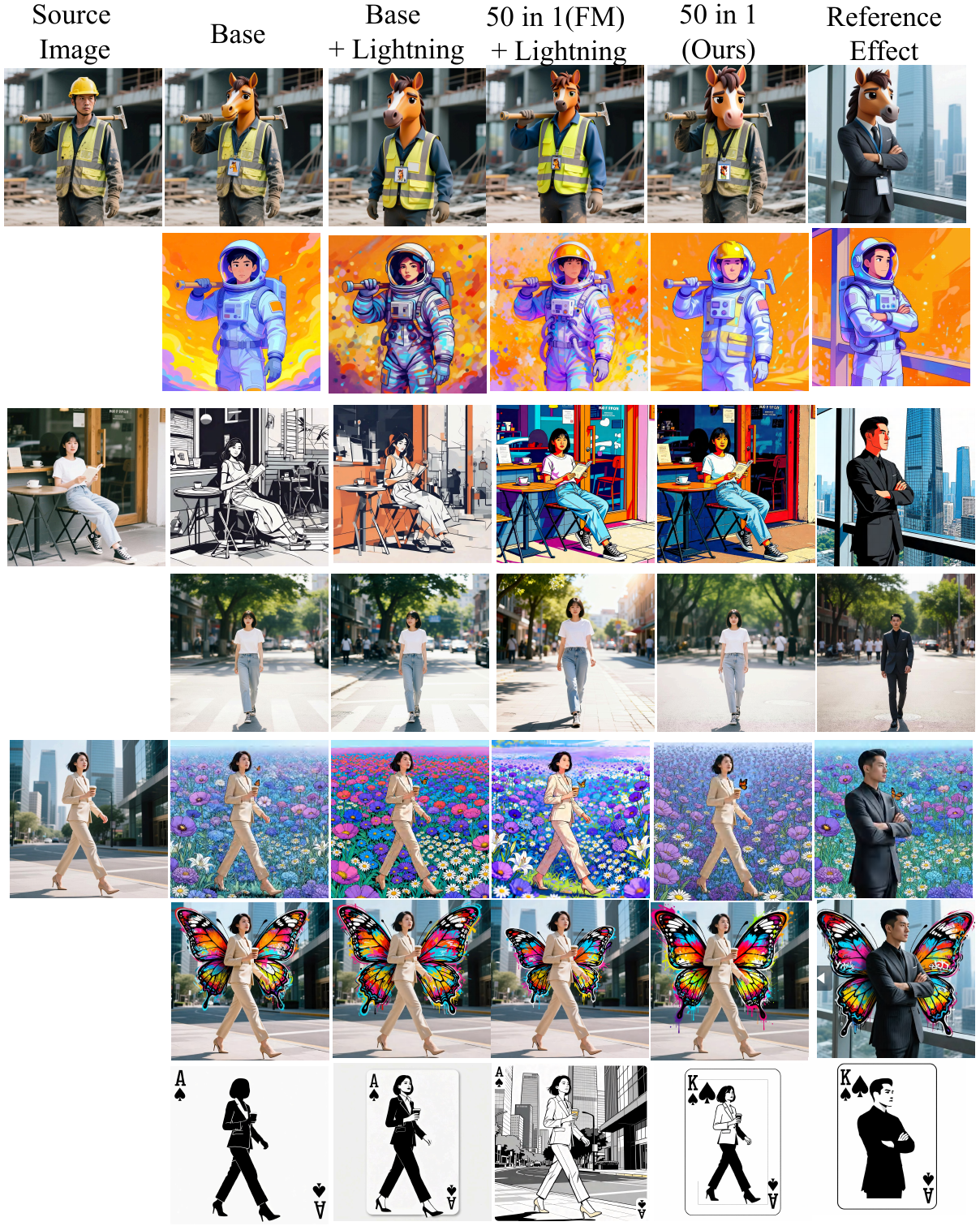}
    \caption{Qualitative Evaluation.}
    \label{fig:visual6}
    \vspace{-0.5cm}
\end{figure*}

We provide additional visualization results in Figure~\ref{fig:visual1},~\ref{fig:visual2},~\ref{fig:visual3},~\ref{fig:visual4},~\ref{fig:visual5} and \ref{fig:visual6}.

\end{document}